%% file: Version_arxiv.tex
\documentclass{article} 
\usepackage{iclr2027_conference,times}
\input{math_commands.tex}

\usepackage{hyperref}
\usepackage{url}
\usepackage{booktabs}
\usepackage{graphicx}
\usepackage{todonotes}
\usepackage{wrapfig}
\usepackage{needspace}
\usepackage{capt-of}

\usepackage{booktabs}
\usepackage{multirow}
\usepackage{graphicx}
\usepackage[table]{xcolor}
\usepackage{tcolorbox}
\usepackage{fontawesome5}
\definecolor{refgray}{gray}{0.55}
\definecolor{CoverRed}{HTML}{A6192E}
\definecolor{CoverBlush}{HTML}{FCF8F8}
\definecolor{PaperLink}{HTML}{245C8B}
\definecolor{HeaderText}{HTML}{535861}
\definecolor{HeaderRule}{HTML}{DCE0E4}
\hypersetup{colorlinks=true,citecolor=PaperLink,linkcolor=PaperLink,urlcolor=PaperLink}

\newcommand{\paperTitle}[1]{PVWander: 3D Memory-guided Panoramic#1Video Wandering with Path-Facing Canonicalization}
\title{\paperTitle{\\}}
\fancypagestyle{paper}{%
  \fancyhf{}
  \fancyhead[C]{\normalfont\fontsize{7}{9}\selectfont\color{HeaderText}\paperTitle{\space}}
  \fancyfoot[C]{\thepage}
  \renewcommand{\headrulewidth}{0.35pt}
  \renewcommand{\headrule}{\hbox to\headwidth{\color{HeaderRule}\leaders\hrule height \headrulewidth\hfill}}
}

\author{
Chuanxin Song\textsuperscript{1,2},\quad Xiaodong Wang\textsuperscript{1,2},\quad
Shijie Chen\textsuperscript{1},\quad Junliang Xie\textsuperscript{3}\\[2pt]
Zhanmi Zhong\textsuperscript{3},\quad Suiying Wu\textsuperscript{3},\quad
Peixi Peng\textsuperscript{1,2,\faEnvelope[regular]}
}
\makeatletter
\renewcommand{\@maketitle}{%
  \par\begingroup
  \centering\sffamily
  {\color{CoverRed}\hrule height 1.2pt}
  \vspace{8pt}
  {\bfseries\fontsize{17}{21}\selectfont\@title\par}
  \vspace{8pt}
  {\color{CoverRed}\hrule height 0.4pt}
  \vspace{10pt}
  {\bfseries\fontsize{11.5}{14}\selectfont\@author\par}
  \vspace{5pt}
  {\fontsize{10.5}{13.5}\selectfont
    \hypersetup{hidelinks}%
    \textsuperscript{1}Peking University\qquad
    \textsuperscript{2}Pengcheng Laboratory\qquad \textsuperscript{3}Frontop\\
    Email: \href{mailto:cxsong25@stu.pku.edu.cn}{cxsong25@stu.pku.edu.cn},\quad
    \href{mailto:pxpeng@pku.edu.cn}{pxpeng@pku.edu.cn}\par}
  \endgroup
}
\makeatother
\renewenvironment{abstract}{%
  \begin{tcolorbox}[colback=CoverBlush,colframe=CoverRed!18,boxrule=0.35pt,
    arc=6pt,outer arc=6pt,boxsep=0pt,left=20pt,right=20pt,top=13pt,bottom=13pt,
    before skip=12pt,after skip=0pt]
  \fontsize{10.5}{13}\selectfont\noindent
  {\sffamily\bfseries\textcolor{CoverRed}{Abstract.}}\enspace\ignorespaces
}{%
  \par\vspace{9pt}
  \noindent{\sffamily\fontsize{10.5}{13}\selectfont
    \faGlobe\enspace\textbf{Project:}\enspace
    \hypersetup{urlcolor=CoverRed}\url{https://chuanxinsong.github.io/pvwander-project-page}}%
  \hfill\raisebox{-0.5\height+0.5ex}{\includegraphics[width=0.43in]{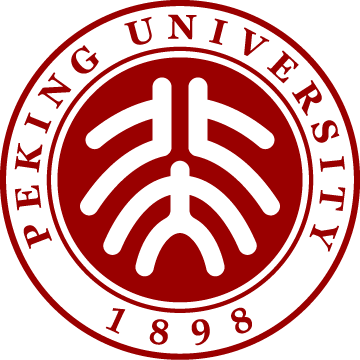}}\par
  \end{tcolorbox}}

\iclrfinalcopy
\begin{document}

\pagestyle{paper}
\thispagestyle{plain}
\enlargethispage{8pt}
\vspace*{-0.72in}
\noindent\makebox[\textwidth][c]{%
\begin{minipage}{\dimexpr\paperwidth-2in\relax}
\maketitle
\vspace{14pt}
{\centering
  \includegraphics[width=\linewidth]{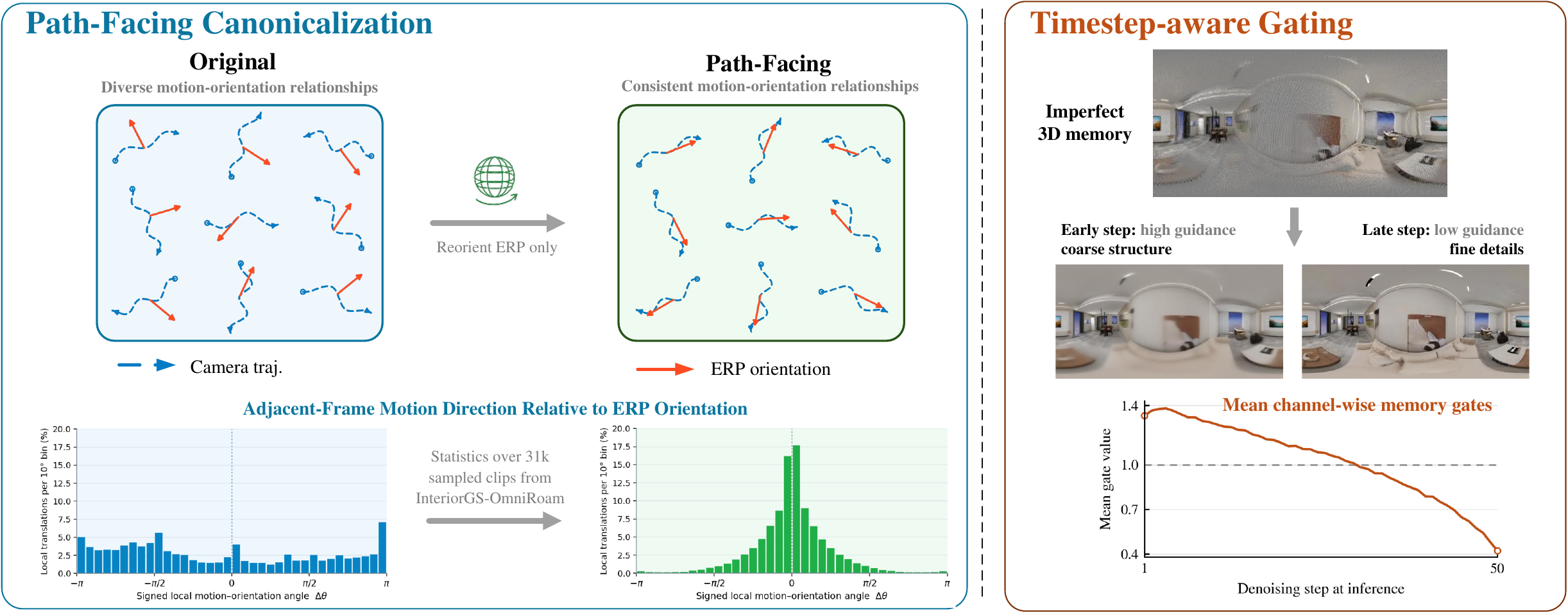}\par}
\begingroup
  \small
  \setlength{\abovecaptionskip}{5pt}
  \captionof{figure}{Two core designs in PVWander. Left: Path-Facing canonicalization reorients ERP to reduce variation in motion direction relative to per-frame ERP orientation.
  Right: Timestep-aware gating modulates reprojected memory features over 50 denoising steps, from high to low noise.}
  \label{fig:intro-teaser}
\endgroup
\begin{abstract}
Long-horizon panoramic video wandering aims to continuously synthesize scene-consistent $360^\circ$ observations along user-specified camera trajectories from an initial equirectangular projection (ERP) panorama. 
Existing panoramic video models have achieved promising results in camera-controlled generation, while recent 3D-guided approaches further provide geometric guidance for improving scene consistency. However, stable long-horizon wandering remains challenging due to diverse motion--orientation relationships and imperfect memory guidance induced by accumulated generation artifacts and depth inaccuracies. 
To address these challenges, we propose \textbf{PVWander}, a 3D-memory-guided framework that accumulates generated observations into an online point-cloud memory and reprojects it to guide subsequent generation, with two complementary designs: \textit{Path-Facing canonicalization} for motion--orientation alignment and \textit{Timestep-aware gating modulation} for adaptive memory guidance.
\textit{Path-Facing canonicalization} exploits the horizontal cyclic symmetry of ERP panoramas to uniformly align each clip's ERP orientation with its principal translation direction, yielding more consistent, forward-like relative motions while preserving the prescribed camera path.
\textit{Timestep-aware gating modulation} exploits the coarse-to-fine nature of diffusion denoising to adaptively regulate reprojected memory guidance through timestep-dependent channel-wise gates, preserving its structural benefits while mitigating visual artifacts from imperfect projections.
Extensive experiments demonstrate that PVWander achieves improved long-horizon generation quality, while ablations validate the effectiveness of both designs.
\end{abstract}
\end{minipage}}\par

\newpage

\renewcommand{\normalsize}{\fontsize{10pt}{13pt}\selectfont}
\normalsize
\section{Introduction}

Building generative models that can continuously unfold a 3D world along camera motion is an important step toward spatial intelligence. Such models can provide controllable virtual worlds for applications such as robotics and virtual reality~\citep{wen3DSceneGeneration2025}. Equirectangular panoramas are naturally suited to this setting by capturing the full 360° surrounding view at each camera location~\citep{yangMatrix3DOmnidirectionalExplorable2025,sunABot3DWorld0Universal2026}. In this work, we focus on \emph{long-horizon panoramic video wandering}: given an initial ERP panorama and a series of user-specified camera trajectories, the goal is to continuously generate panoramic observations along these trajectories. 

Existing camera-controlled panoramic video generation methods have achieved promising results by synthesizing panoramic observations under user-specified camera trajectories~\citep{jiCamPVGCameraControlledPanoramic2025,yinPanoWorldXGeneratingExplorable2025}. 
However, maintaining scene appearance and structure becomes challenging when these methods are recursively applied over longer trajectories.
Recent approaches~\citep{yangMatrix3DOmnidirectionalExplorable2025,chenPantheon360TamingDigital,wangRein3DReinforced3D2026} introduce explicit 3D representations (e.g., point clouds or 3D Gaussians) derived from the input panorama to provide geometric guidance.
Yet these methods primarily focus on single-segment generation, leaving closed-loop integration of newly generated observations during long-horizon rollout unexplored.
In perspective video domain, recent works~\citep{wuVideoWorldModels2025,shenLyra20Explorable2026} instead maintain an online 3D memory that is continuously updated and reprojected to condition subsequent video segments.
Building on this paradigm, we propose \textbf{PVWander}, a 3D-memory-guided framework that continuously incorporates newly generated panoramic observations to guide long-horizon wandering.

Even with online 3D memory, stable long-horizon panoramic video generation still faces two key challenges.
First, free-form camera control involves highly varying motion directions relative to the viewing orientation, ranging from forward/backward to lateral motion. Such diverse motion--orientation relationships can exacerbate visual artifacts and temporal drift over long trajectories. 
Second, while 3D memory provides useful geometric guidance through reprojection, it is reconstructed from generated observations and estimated depth. Generation artifacts and depth inaccuracies can therefore introduce unreliable local details into reprojected views, potentially propagating errors across segments.
To address these challenges, we introduce two complementary designs: \textit{Path-Facing canonicalization} and \textit{Timestep-aware gating modulation}, as illustrated in Fig.~\ref{fig:intro-teaser}.

First, to reduce variation in motion--orientation relationships, we exploit the horizontal cyclic symmetry of ERP panoramas, which allows their orientation to change without altering scene content (see Fig.~\ref{fig:erp-rotation} in Appendix~\ref{sec:appendix-panorama-conversion}). Based on this observation, we propose \textit{Path-Facing canonicalization}, which estimates the principal translation direction of each segment and aligns all-frame ERP orientation with this direction. As illustrated in Fig.~\ref{fig:intro-teaser} (left), this reorientation transforms diverse motion--orientation relationships into a more consistent forward-like configuration while preserving the prescribed camera path.
Second, to better exploit imperfect 3D memory projections, we aim to retain their useful coarse geometric guidance while reducing the influence of unreliable local details. Diffusion denoising generally follows a coarse-to-fine process, establishing global structure before refining local appearance~\citep{qianBoostingDiffusionModels2024,renFrequencyAwareFlowMatching2026}. This motivates our \textit{Timestep-aware gating modulation}, which predicts timestep-dependent channel-wise gates to adapt memory guidance to different denoising stages. As shown in Fig.~\ref{fig:intro-teaser} (right), the learned mean gate is stronger early in denoising and generally decreases during later refinement.

Extensive experiments evaluate PVWander on long-horizon generation and zero-shot generalization, while ablations demonstrate the effectiveness of its key designs.
Our main contributions are threefold:
(1) We propose \textbf{PVWander}, a 3D-memory-guided closed-loop framework for long-horizon panoramic video wandering that combines motion--orientation canonicalization with adaptive memory guidance.
(2) We propose \textit{Path-Facing canonicalization}, which exploits ERP cyclic symmetry to reorient panoramas along the principal translation direction, yielding more consistent, forward-like motion--orientation relationships.
(3) We propose \textit{Timestep-aware gating modulation}, which adaptively modulates imperfect 3D memory guidance through timestep-dependent channel-wise gates.

\section{Related Work}
\label{sec:related-work}
\label{sec:appendix-related-work}

\paragraph{Panoramic Video Generation.}
Panoramic video generation has been explored from both content synthesis and controllable exploration perspectives. For content synthesis, existing methods extend pretrained text-to-video models with panorama-specific designs to generate immersive videos with a full $360^\circ$ field of view~\citep{wang360DVDControllablePanorama2024,xiaPanoWanLiftingDiffusion2025,zhangPanoDiTPanoramicVideos2025}. 
For controllable exploration, camera-controlled approaches synthesize panoramic videos along specified camera trajectories~\citep{jiCamPVGCameraControlledPanoramic2025,yinPanoWorldXGeneratingExplorable2025}. Several methods further simplify trajectory conditioning by decoupling camera rotation or using translation-only motion~\citep{zhangPanFlowDecoupledMotion2025,cheng360ExplorerExploring4D2026,liuOmniRoamWorldWandering2026}. However, regardless of whether these simplifications are applied, free-form translations can exhibit highly diverse directions relative to the ERP orientation. In contrast, our \textit{Path-Facing canonicalization} exploits the horizontal cyclic symmetry of panoramas to align each segment's ERP orientation with its principal translation direction, yielding more consistent, forward-like motion--orientation relationships.
Explicit 3D representations have also been introduced to provide geometric guidance using meshes, point clouds, or 3D Gaussians~\citep{yangMatrix3DOmnidirectionalExplorable2025,chenPantheon360TamingDigital,wangRein3DReinforced3D2026,sunABot3DWorld0Universal2026}. However, most do not maintain a persistent 3D memory that evolves with newly generated observations, limiting their ability to continuously accumulate scene information over long horizons. 
Similar to our formulation, EvoWorld~\citep{wangEvoWorldEvolvingPanoramic2025} maintains an evolving 3D point-cloud memory that is iteratively updated and reprojected to guide subsequent panoramic generation. 
However, repeated memory updates can also accumulate generation artifacts and depth estimation errors.
PVWander further emphasizes robustness to these accumulated errors through \textit{Timestep-aware gating modulation}.

\paragraph{Camera-Controlled Long-Horizon Video Generation.}
Camera-controlled video diffusion models inject camera trajectories through pose parameters or dense ray embeddings~\citep{wangMotionCtrlUnifiedFlexible2024,heCameraCtrlEnablingCamera2024}. Recent methods further incorporate rendered 3D guidance to strengthen geometric control~\citep{yuViewCrafterTamingVideo2025,renGEN3C3DInformedWorldConsistent2025}. However, most models operate within a fixed temporal context, which can compromise long-horizon consistency when earlier observations fall outside the model's effective context window.
Memory-aware methods address this limitation by retrieving relevant historical views~\citep{yuContextMemorySceneConsistent2025,liVMemConsistentInteractive2025,liPanoWorldRealworldPanoramic2026} or accumulating observations into persistent 3D scene representations that can be reprojected to future viewpoints~\citep{wuVideoWorldModels2025,wangEvoWorldEvolvingPanoramic2025}. AnchorWeave~\citep{wang2026anchorweave} and Lyra 2.0~\citep{shenLyra20Explorable2026} further combine geometry-aware retrieval with per-frame 3D representations to establish correspondences with relevant historical observations. 
Geometry-grounded memories improve long-horizon consistency, but continuously updating them from generated observations can introduce a closed-loop error-propagation problem, as visual artifacts and depth inaccuracies may be repeatedly written into the memory and affect subsequent predictions. We therefore introduce \textit{Timestep-aware gating modulation}, which exploits the coarse-to-fine nature of diffusion denoising~\citep{renFrequencyAwareFlowMatching2026} to adaptively regulate memory guidance with timestep-dependent gates, preserving its benefits for structural consistency while mitigating visual artifacts.

\section{Method}
\label{sec:method}

\begin{figure}[t]
\centering
\includegraphics[width=\linewidth]{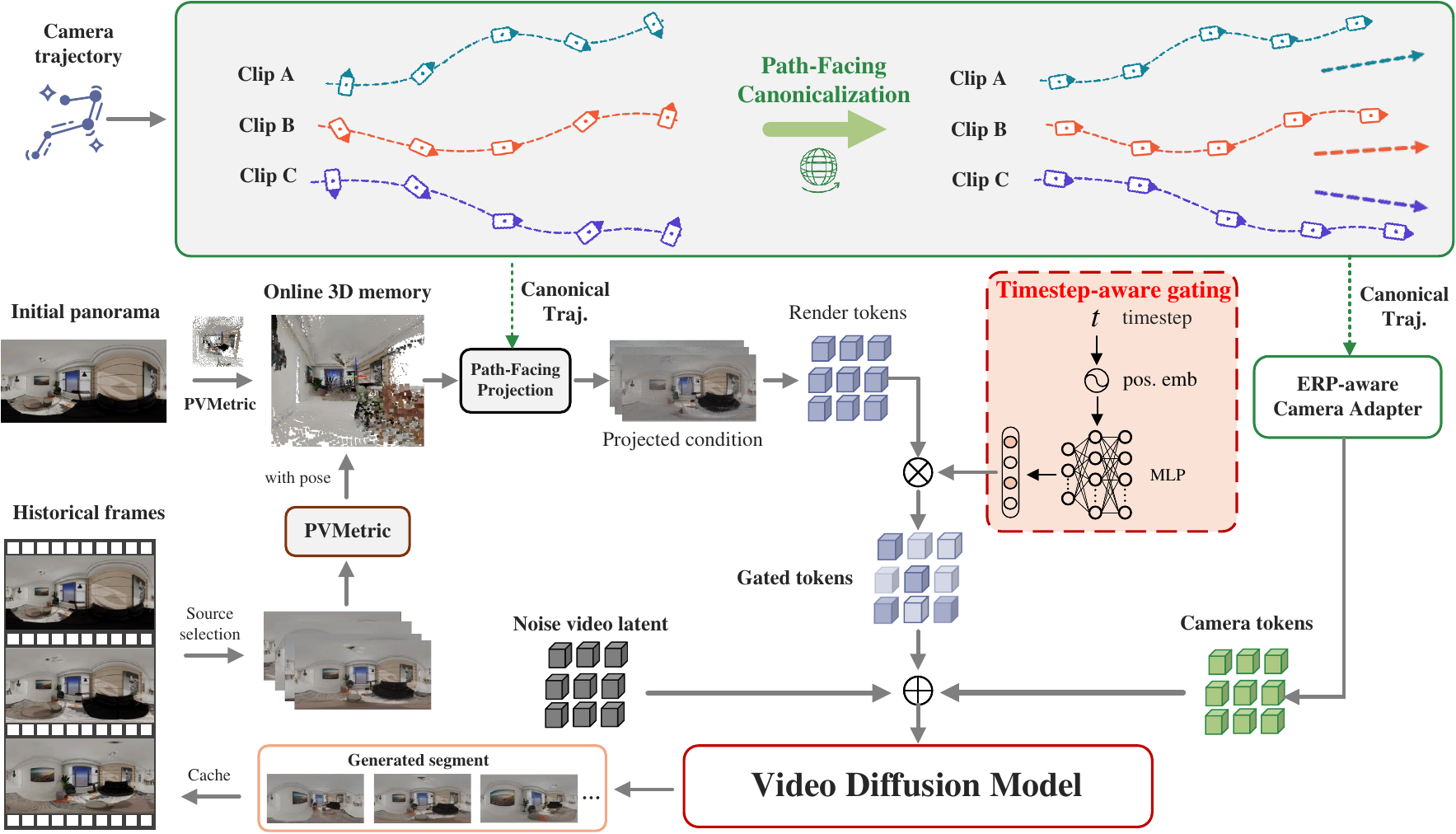}
\caption{Overview of PVWander. Online 3D memory accumulates and reprojects generated observations to guide subsequent segments. Path-Facing canonicalization aligns ERP orientation with the principal translation direction for more consistent motion--orientation relationships, while Timestep-aware gating balances useful geometric guidance and unreliable local details across denoising stages.}
\label{fig:pvwander-framework}
\end{figure}

Given an initial ERP panorama and user-specified long camera trajectories, \textbf{PVWander} generates panoramic video segment by segment with an online 3D point-cloud memory. As shown in Fig.~\ref{fig:pvwander-framework}, the memory is reprojected to target viewpoints to guide each segment, and newly generated observations are lifted into 3D to update the memory for subsequent generation.
We first describe this memory-guided generation framework, followed by \textit{Path-Facing canonicalization} for motion--orientation alignment and \textit{Timestep-aware gating modulation} for adaptive guidance from imperfect memory projections. Preliminaries on the video diffusion backbone and 3D memory-guided generation are provided in Appendix~\ref{sec:appendix-preliminaries}.

\subsection{Online 3D Memory-guided Panoramic Generation}

\paragraph{Online 3D Memory.}
We represent the memory at generation step $s$ as a set of colored 3D points $\mathcal{M}^{(s)}=\{(\mathbf{p}_j,\mathbf{c}_j)\}_{j=1}^{N_s}$, where $\mathbf{p}_j\in\mathbb{R}^3$ denotes a world-coordinate point and $\mathbf{c}_j\in\mathbb{R}^3$ stores its RGB observation. The memory is initialized by lifting the input panorama into 3D using metric radial depth estimated by PVMetric (see Appendix~\ref{sec:appendix-pvmetric}). After each segment, the generated observations are converted into 3D using their 
estimated metric depth, back-projected into the common world coordinate system, 
and added to the memory.

Before generating the next segment, following~\citep{shenLyra20Explorable2026}, we select source observations based on their geometric coverage of target viewpoints, favoring complementary views that cover previously uncovered regions. The memory points associated with the selected observations are transformed into each target camera coordinate system and projected onto the ERP image plane, producing partial panoramic renders that provide spatial guidance for subsequent generation.

\paragraph{Memory-Guided Video Generation.}
Each segment is conditioned on its first frame, the target camera trajectory, and the reprojected memory renders. The first-frame condition is the input panorama for the initial segment and the final frame of the preceding segment thereafter.
For camera conditioning, following~\citep{jiCamPVGCameraControlledPanoramic2025}, we adapt Pl\"ucker ray encoding to ERP geometry and encode the ray maps using the existing camera adapter of the Wan backbone (see Appendix~\ref{sec:appendix-erp-camera} for details).
For memory guidance, we encode the panoramic renders with the video VAE and map the resulting latents, together with a projection-validity mask indicating memory coverage, into render tokens using a lightweight render adapter.
Following Matrix-3D~\citep{yangMatrix3DOmnidirectionalExplorable2025}, we add the camera and render tokens to the patch-embedded video features before passing them to the diffusion backbone:
\begin{equation}
\mathbf{h}_t
=
\mathbf{h}^{\mathrm{base}}_t
+
\mathbf{h}^{\mathrm{cam}}
+
\mathbf{h}^{\mathrm{render}},
\end{equation}
where $t$ denotes the diffusion timestep, $\mathbf{h}^{\mathrm{base}}_t$ denotes the patch-embedded features of the noisy video latent together with the first-frame condition, and $\mathbf{h}^{\mathrm{cam}}$ and $\mathbf{h}^{\mathrm{render}}$ denote the camera and memory-render conditioning tokens, respectively.

While this closed-loop process enables scene information to be reused across segments, stable long-horizon generation still faces two challenges. 
First, free-form trajectories induce diverse motion directions relative to ERP orientation, increasing the difficulty of motion modeling and exacerbating visual degradation over long-horizon generation.
Second, the online memory may accumulate generation artifacts and depth inaccuracies, which can propagate through subsequent reprojections. 
We address these issues with \textit{Path-Facing canonicalization} and \textit{Timestep-aware gating modulation}, respectively.

\subsection{Path-Facing Canonicalization}
\label{sec:path-facing}

Path-Facing canonicalization reduces motion--orientation variation by aligning each segment's ERP orientation with its principal translation direction. It exploits the horizontal cyclic symmetry of ERP panoramas, which allows yaw reorientation through circular shifts while preserving scene coverage (see Fig.~\ref{fig:erp-rotation} in Appendix~\ref{sec:appendix-panorama-conversion}).

Given a target trajectory segment with horizontal camera coordinates $\{\mathbf{q}_i\}_{i=1}^{L}$, we apply Principal Component Analysis (PCA) to estimate its principal translation axis:
\begin{equation}
\mathbf{v}
=
\operatorname{PCA}_1
\left(
\{\mathbf{q}_i-\bar{\mathbf{q}}\}_{i=1}^{L}
\right),
\end{equation}
where $\bar{\mathbf{q}}$ denotes the mean camera position. We orient this axis according to the temporal progression of the trajectory and use the resulting direction $\mathbf{v}$ as the shared ERP orientation for all frames in the segment, while leaving camera positions unchanged.
As shown in Fig.~\ref{fig:pvwander-framework}, we reorient the first-frame condition and use the aligned camera poses for both memory reprojection and camera conditioning. Generated observations are then back-projected using these same poses, keeping the memory in a consistent world coordinate system.
The generated panoramas can be rotated back to their original orientations if needed.

As shown by the statistics in Fig.~\ref{fig:intro-teaser} (left), Path-Facing makes frame-wise motion directions relative to the ERP orientation more concentrated around the forward direction. During training, we primarily use Path-Facing while retaining orientation augmentation for robustness. 
It requires no additional learnable parameters or architectural changes and incurs negligible computational overhead, allowing direct application to existing panoramic video generators.

\subsection{Timestep-aware gating Modulation}
\label{sec:timestep-gating}

Reprojected memory may contain unreliable local details caused by generation artifacts and depth inaccuracies, while still retaining useful coarse spatial structure. This motivates us to adapt its guidance to the coarse-to-fine denoising process: early high-noise stages primarily establish global structure, whereas later stages refine local appearance~\citep{qianBoostingDiffusionModels2024,renFrequencyAwareFlowMatching2026}.

Simply disabling memory guidance during later stages may also discard useful information and applies the same cutoff to all feature channels (see Appendix~\ref{sec:appendix-hard-cutoff}). We therefore introduce \textit{Timestep-aware gating modulation}, which learns channel-wise gates conditioned on the denoising timestep, allowing memory features to be attenuated or amplified without a hard cutoff.

As shown in Fig.~\ref{fig:pvwander-framework}, we encode the current timestep $t$ with a sinusoidal embedding $\mathbf{e}_t=\operatorname{PE}(t)$. A lightweight two-layer MLP with a SiLU activation predicts channel-wise gates that modulate the render tokens before fusion with the diffusion backbone:
\begin{equation}
\mathbf{g}_t = \mathbf{1} + \tanh\left(f_{\mathrm{gate}}(\mathbf{e}_t)\right),
\qquad
\mathbf{h}_t = \mathbf{h}^{\mathrm{base}}_t + \mathbf{h}^{\mathrm{cam}} + \mathbf{g}_t\odot\mathbf{h}^{\mathrm{render}}.
\end{equation}
where $f_{\mathrm{gate}}$ denotes the timestep-conditioned MLP and $\mathbf{g}_t\in(0,2)^C$, with $C$ denoting the render token dimension. We zero-initialize the MLP's output layer so that $\mathbf{g}_t=\mathbf{1}$ at initialization, preserving the original render conditioning.

As shown in Fig.~\ref{fig:gating-timestep}, the learned mean gate is stronger early in denoising and generally decreases during later refinement, consistent with retaining structural guidance while reducing reliance on imperfect local details.

\section{Experiments}

\subsection{Experimental Settings}
\label{sec:experi_setting}

We conduct main experiments on InteriorGS-OmniRoam~\citep{liuOmniRoamWorldWandering2026}, following its split of 876 training and 98 evaluation scenes. 
Since ERP panoramas provide full $360^\circ$ coverage, the dataset uses a fixed ERP orientation across its diverse camera trajectories. 
Evaluation covers both the first 81-frame segment and the full eight-segment (641-frame) rollout using SSIM, PSNR, LPIPS, FID, and cubemap-based $\mathrm{FID}_{\mathrm{cube}}$. Trajectory fidelity is measured by TransErr and RotErr computed from poses estimated by ViPE~\citep{huangViPEVideoPose2025}. 
We further evaluate zero-shot generalization on 90 RealSee3D scenes~\citep{liArgusMetricPanoramic2026} using 481-frame closed-loop trajectories. SSIM-Loop, PSNR-Loop, and LPIPS-Loop compare the initial and final panoramas, while trajectory fidelity is evaluated over the full sequence.

For our main experiments, we initialize PVWander from \textit{Wan2.1-Fun-V1.1-1.3B-Control-Camera} and train it on 81-frame clips using LoRA and progressively increasing resolutions of $640\times320$, $960\times480$, and $1440\times720$. We use PVMetric depth for memory construction during both training and inference.

For a fair comparison with EvoWorld~\citep{wangEvoWorldEvolvingPanoramic2025}, we train an SVD-based variant of PVWander on the same Unity\_Curve dataset. We use EvoWorld's train--test split of 3,666 training episodes and 200 test episodes. Following its evaluation protocol, we report results for both the first segment (25 frames) and the full three-segment rollout (73 frames).

Detailed evaluation protocols and implementation settings are provided in Appendix~\ref{sec:appendix-experimental-settings}.

\subsection{Comparison with State-of-the-Art Methods}
We compare PVWander with the ERP-adapted Wan2.1-Fun~\citep{wanWanOpenAdvanced2025} baseline and recent representative panoramic video wandering methods, including Matrix-3D~\citep{yangMatrix3DOmnidirectionalExplorable2025}, OmniRoam~\citep{liuOmniRoamWorldWandering2026}, PanoWorld~\citep{liPanoWorldRealworldPanoramic2026}, and EvoWorld~\citep{wangEvoWorldEvolvingPanoramic2025} (evaluated separately on Unity\_Curve).
For methods originally designed for single-segment generation, we extend them to long-horizon wandering through autoregressive rollout, using the last frame of each segment to initialize the next.
Since the full OmniRoam pipeline follows a fixed 8-segment preview-and-refine protocol, we also evaluate its preview model, denoted as \textit{OmniRoam-P}, under the same autoregressive wandering setting as the other methods.

\paragraph{Quantitative Comparison}

\begin{table*}[t]
    \centering
    \caption{Quantitative comparison on InteriorGS-OmniRoam for the first segment and the full 8-segment rollout. Best results at each resolution are shown in bold. 
    Gray entries are reference results: the ground-truth (GT) row reports ViPE trajectory estimates from GT frames, and OmniRoam-P$_{s=8}$ denotes the single-pass stride-8 sparse preview.
    Matrix-3D$_{\mathrm{adapted}}^{\mathrm{MoGe}}$ uses GT depth for training and GT-aligned MoGe depth at inference, while Matrix-3D$_{\mathrm{adapted}}^{\mathrm{PVMetric}}$ uses PVMetric consistently during both training and inference.}
    \label{tab:main-comparison}
    \resizebox{\textwidth}{!}{
    \begin{tabular}{lllcccccccccccccc}
        \toprule
        \multirow{2}{*}{Resolution} & \multirow{2}{*}{Methods} & \multirow{2}{*}{Backbone}
        & \multicolumn{7}{c}{First Segment (81 frames)}
        & \multicolumn{7}{c}{8 Segments (641 frames)} \\
        \cmidrule(lr){4-10} \cmidrule(lr){11-17}
        & & & SSIM$\uparrow$ & PSNR$\uparrow$ & LPIPS$\downarrow$ & FID$\downarrow$ & $\mathrm{FID}_{\mathrm{cube}}\downarrow$ & TransErr$\downarrow$ & RotErr$\downarrow$ & SSIM$\uparrow$ & PSNR$\uparrow$ & LPIPS$\downarrow$ & FID$\downarrow$ & $\mathrm{FID}_{\mathrm{cube}}\downarrow$ & TransErr$\downarrow$ & RotErr$\downarrow$ \\
        \midrule

        \multirow{4}{*}{$1440\times720$}
        & \textcolor{gray}{GT} & \textcolor{gray}{--} & \textcolor{gray}{--} & \textcolor{gray}{--} & \textcolor{gray}{--} & \textcolor{gray}{--} & \textcolor{gray}{--} & \textcolor{gray}{0.0017} & \textcolor{gray}{0.1043} & \textcolor{gray}{--} & \textcolor{gray}{--} & \textcolor{gray}{--} & \textcolor{gray}{--} & \textcolor{gray}{--} & \textcolor{gray}{0.0162} & \textcolor{gray}{0.3502} \\
        & OmniRoam & Wan 1.3B & 0.6930 & 17.35 & 0.3217 & 22.46 & 39.75 & 0.1169 & 0.7338 & 0.6063 & 14.59 & 0.4662 & \textbf{19.88} & 38.38 & 0.2869 & 1.7781 \\
        & Matrix-3D & Wan 5B & 0.7404 & 19.36 & 0.2709 & 23.22 & 45.85 & 0.0143 & 0.5015 & 0.5748 & 12.76 & 0.5765 & 82.49 & 100.76 & 1.1411 & 9.4044 \\
        & PVWander & Wan 1.3B & \textbf{0.7935} & \textbf{21.62} & \textbf{0.1750} & \textbf{16.18} & \textbf{27.15} & \textbf{0.0099} & \textbf{0.2513} & \textbf{0.6810} & \textbf{16.32} & \textbf{0.3983} & 23.65 & \textbf{36.92} & \textbf{0.1890} & \textbf{1.1388} \\

        \midrule
        \multirow{4}{*}{$960\times480$}
        & PanoWorld & Wan 5B & 0.6000 & 15.73 & 0.3952 & 28.28 & 43.03 & 0.1390 & 0.3768 & 0.4546 & 11.96 & 0.5913 & 26.17 & 50.38 & 0.7299 & 1.5944 \\
        & OmniRoam-P$_{s=1}$ & Wan 1.3B & 0.6823 & 18.31 & 0.2536 & 19.31 & 30.53 & 0.0369 & 0.4757 & 0.5383 & 13.48 & 0.4994 & 32.57 & 48.15 & 0.3794 & 1.7844 \\
        & \textcolor{gray}{OmniRoam-P$_{s=8}$} & \textcolor{gray}{Wan1.3B} & \textcolor{gray}{--} & \textcolor{gray}{--} & \textcolor{gray}{--} & \textcolor{gray}{--} & \textcolor{gray}{--} & \textcolor{gray}{--} & \textcolor{gray}{--} & \textcolor{gray}{0.5823} & \textcolor{gray}{15.02} & \textcolor{gray}{0.4392} & \textcolor{gray}{23.14} & \textcolor{gray}{55.87} & \textcolor{gray}{0.2546} & \textcolor{gray}{1.5701} \\
        & PVWander & Wan 1.3B & \textbf{0.7777} & \textbf{21.92} & \textbf{0.1656} & \textbf{16.79} & \textbf{29.29} & \textbf{0.0124} & \textbf{0.2789} & \textbf{0.6467} & \textbf{16.44} & \textbf{0.4038} & \textbf{23.07} & \textbf{37.56} & \textbf{0.1770} & \textbf{0.9859} \\

        \midrule
        \multirow{4}{*}{$640\times320$}
        & Wan2.1-Fun & Wan 1.3B & 0.7181 & 20.88 & 0.1717 & 18.78 & 34.98 & 0.0225 & 0.4612 & 0.5492 & 15.36 & 0.4220 & 24.82 & 45.21 & \textbf{0.2267} & 1.5698 \\
        & Matrix-3D$_{\mathrm{adapted}}^{\mathrm{MoGe}}$ & Wan 1.3B & 0.7087 & 20.48 & 0.1838 & 18.84 & 33.77 & \textbf{0.0146} & \textbf{0.3212} & 0.5515 & 14.85 & 0.4732 & 29.39 & 47.09 & 0.6174 & 2.2268 \\
        & Matrix-3D$_{\mathrm{adapted}}^{\mathrm{PVMetric}}$ & Wan 1.3B & 0.7301 & 21.29 & 0.1686 & 18.95 & 35.50 & 0.0195 & 0.3685 & 0.5526 & 15.33 & 0.4304 & 27.02 & 47.69 & 0.2911 & 1.6191 \\
        & PVWander & Wan 1.3B & \textbf{0.7353} & \textbf{21.44} & \textbf{0.1615} & \textbf{17.92} & \textbf{33.08} & 0.0194 & 0.3532 & \textbf{0.5867} & \textbf{16.25} & \textbf{0.4032} & \textbf{23.75} & \textbf{38.13} & 0.3521 & \textbf{1.3363} \\

        \bottomrule
    \end{tabular}
    }
\end{table*}

As reported in Tab.~\ref{tab:main-comparison}, across all three resolutions, PVWander achieves the strongest overall appearance fidelity, consistently obtaining higher SSIM and PSNR and lower LPIPS than the competing methods for both the first segment and the full rollout. The advantage remains substantial after 641 frames, while trajectory fidelity remains competitive across resolutions. 
Appendix~\ref{sec:appendix-computational-cost} further compares the computational cost and runtime of these methods.

\begin{wrapfigure}{r}{0.4\textwidth}
    \vspace{-\baselineskip}
    \centering
    \includegraphics[width=\linewidth]{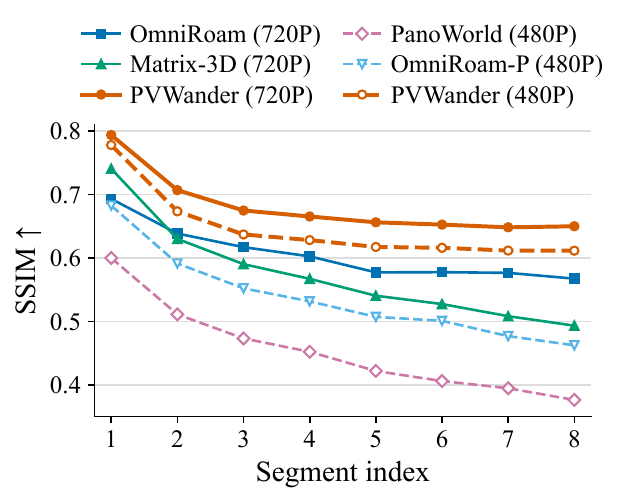}
    \caption{Segment-wise SSIM over the eight-segment rollout.}
    \label{fig:segment-ssim}
\end{wrapfigure}

Fig.~\ref{fig:segment-ssim} further examines how generation quality evolves throughout the rollout. PVWander maintains the highest SSIM among the compared methods at the corresponding resolution across all eight segments. Although generation quality decreases as the rollout proceeds, PVWander remains relatively stable in the later stages. In comparison, Matrix-3D at $1440\times720$, as well as PanoWorld and OmniRoam-P at $960\times480$, exhibit more pronounced degradation. 

OmniRoam also achieves strong long-horizon performance with its full preview-and-refinement pipeline, where sparse frames spanning the entire 8-segment trajectory guide segment-wise refinement. When the same preview model is rolled out autoregressively at stride 1 as \textit{OmniRoam-P}, performance degrades more noticeably. In contrast, its single-pass stride-8 preview remains relatively stable, suggesting that preview guidance spanning the full trajectory contributes to the pipeline's long-horizon performance.

Matrix-3D performs strongly on the first segment but degrades considerably during recursive long-horizon generation. Unlike PVWander, it reconstructs geometric guidance independently from each segment's initial frame without accumulating generated observations into a persistent scene memory. To reduce differences in backbone and training data, we further adapt Matrix-3D to our ERP Wan2.1-Fun backbone at $640\times320$. We evaluate two depth settings: Matrix-3D$_{\mathrm{adapted}}^{\mathrm{MoGe}}$ follows the original GT-training/MoGe-inference protocol with GT-aligned depth at inference, while Matrix-3D$_{\mathrm{adapted}}^{\mathrm{PVMetric}}$ uses PVMetric consistently during both training and inference without GT alignment, matching the depth protocol of PVWander. 
The latter improves first-segment appearance fidelity, indicating that reducing the train--inference depth discrepancy is beneficial. Nevertheless, after the full 8-segment rollout, both variants remain clearly below PVWander in appearance fidelity. These results suggest that matching the training and inference depth sources alone does not close the long-horizon performance gap. Detailed depth and alignment settings are provided in Appendix~\ref{sec:appendix-baselines}.

\begin{figure}[ht]
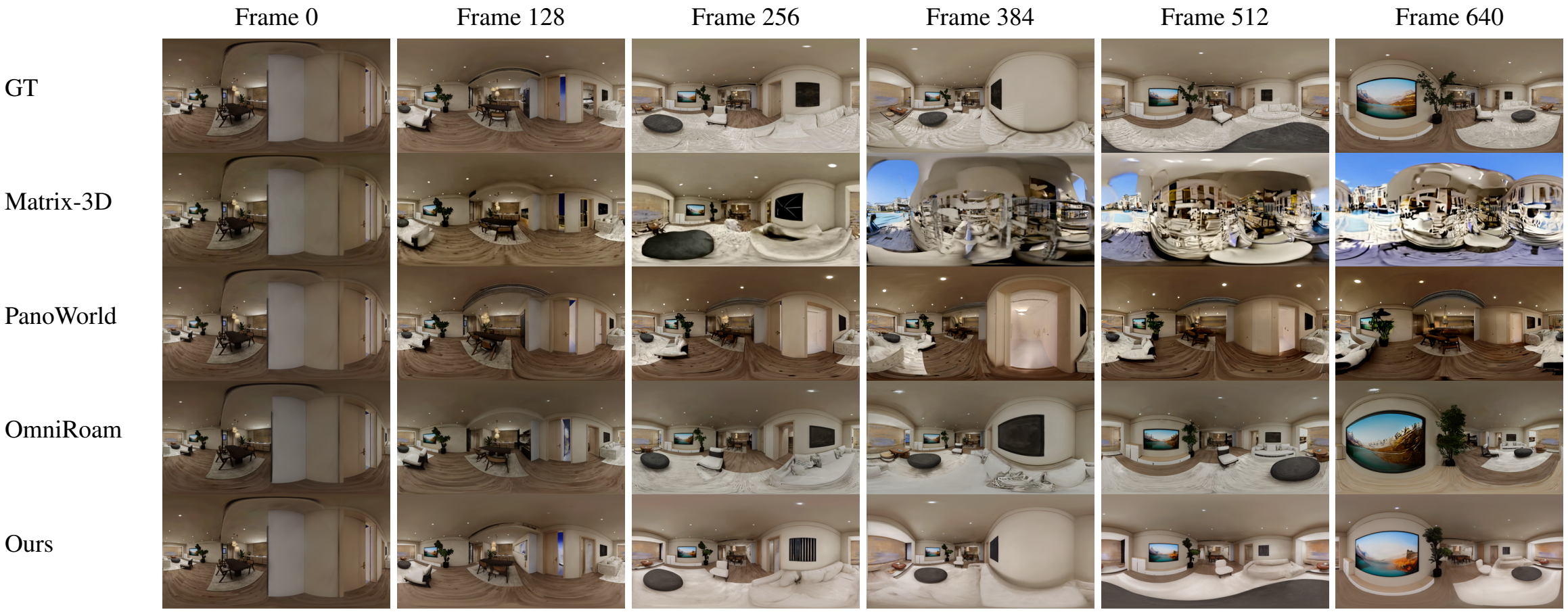

    \centering
    \begingroup
    \setlength{\tabcolsep}{1pt}
    \renewcommand{\arraystretch}{0.8}
    \newcommand{\qualiframe}[2]{\raisebox{-0.5\height}{\includegraphics[width=\dimexpr(\textwidth-54pt)/6\relax]{Fig/quali_com_0900_841496/#1/#2.png}}}
    \scriptsize
    \begin{tabular}{@{}p{42pt}cccccc@{}}
        & Frame 0 & Frame 128 & Frame 256 & Frame 384 & Frame 512 & Frame 640 \\[2pt]
        GT & \qualiframe{GT}{0} & \qualiframe{GT}{128} & \qualiframe{GT}{256} & \qualiframe{GT}{384} & \qualiframe{GT}{512} & \qualiframe{GT}{640} \\[2pt]
        Matrix-3D & \qualiframe{Matrix-3D}{0} & \qualiframe{Matrix-3D}{128} & \qualiframe{Matrix-3D}{256} & \qualiframe{Matrix-3D}{384} & \qualiframe{Matrix-3D}{512} & \qualiframe{Matrix-3D}{640} \\[2pt]
        PanoWorld & \qualiframe{PanoWorld}{0} & \qualiframe{PanoWorld}{128} & \qualiframe{PanoWorld}{256} & \qualiframe{PanoWorld}{384} & \qualiframe{PanoWorld}{512} & \qualiframe{PanoWorld}{640} \\[2pt]
        OmniRoam & \qualiframe{OmniRoam}{0} & \qualiframe{OmniRoam}{128} & \qualiframe{OmniRoam}{256} & \qualiframe{OmniRoam}{384} & \qualiframe{OmniRoam}{512} & \qualiframe{OmniRoam}{640} \\[2pt]
        Ours & \qualiframe{Ours}{0} & \qualiframe{Ours}{128} & \qualiframe{Ours}{256} & \qualiframe{Ours}{384} & \qualiframe{Ours}{512} & \qualiframe{Ours}{640}
    \end{tabular}
    \endgroup
    \caption{Qualitative comparison on InteriorGS-OmniRoam, showing frames sampled every 128 frames throughout the 641-frame rollout. Rows show GT, Matrix-3D, PanoWorld, OmniRoam, and PVWander (Ours).}
    \label{fig:qualitative-comparison}
\end{figure}

\paragraph{Qualitative Comparison.}
Fig.~\ref{fig:qualitative-comparison} compares panoramic frames at fixed intervals throughout the 641-frame rollout. 
Matrix-3D produces plausible results in the early stages but gradually develops severe structural distortions and visual artifacts under recursive generation, consistent with its quantitative degradation. 
PanoWorld maintains reasonable visual quality, but its generated views increasingly deviate from the corresponding GT viewpoints, indicating weaker trajectory adherence in long-horizon evaluation. Notably, the released PanoWorld model supports only single-segment generation and does not expose its full long-range memory mechanism; we therefore extend it autoregressively using the last frame of each segment, which likely contributes to its degradation over extended rollout. 
OmniRoam better preserves the overall scene layout, but exhibits discrepancies in the relative scale of scene elements. 
In contrast, PVWander maintains scene structure and object proportions closer to the GT throughout the trajectory, consistent with the spatial constraints provided by its continuously updated 3D memory.

\begin{table}[ht]
    \centering
    \caption{Zero-shot evaluation on 90 RealSee3D scenes with 481-frame closed-loop trajectories. Loop metrics compare the final generated panorama with the initial panorama, while TransErr and RotErr measure trajectory fidelity over all 481 frames. Best results at each resolution are shown in bold.}
    \label{tab:zeroshot-realsee3d}
    \resizebox{0.8\textwidth}{!}{
    \begin{tabular}{llccccc}
        \toprule
        Method & Resolution & SSIM-Loop$\uparrow$ & PSNR-Loop$\uparrow$ & LPIPS-Loop$\downarrow$ & TransErr$\downarrow$ & RotErr$\downarrow$ \\
        \midrule
        OmniRoam & $1440\times720$ & 0.6851 & 18.30 & \textbf{0.3030} & 0.1722 & 0.9942 \\
        Matrix-3D & $1440\times720$ & 0.6824 & 13.77 & 0.5390 & 0.3935 & 3.0133 \\
        PVWander & $1440\times720$ & \textbf{0.7821} & \textbf{19.56} & 0.3064 & \textbf{0.0477} & \textbf{0.6057} \\
        \midrule
        PanoWorld & $960\times480$ & 0.5447 & 14.59 & 0.4832 & 0.4055 & 0.8093 \\
        Matrix-3D & $960\times480$ & 0.6427 & 13.32 & 0.6022 & 0.4059 & 0.9482 \\
        OmniRoam-P & $960\times480$ & 0.5064 & 12.05 & 0.5838 & 0.6125 & 2.2802 \\
        PVWander & $960\times480$ & \textbf{0.7550} & \textbf{18.83} & \textbf{0.3314} & \textbf{0.05859} & \textbf{0.6906} \\
        \bottomrule
    \end{tabular}
    }
\end{table}

\begin{figure}[ht]
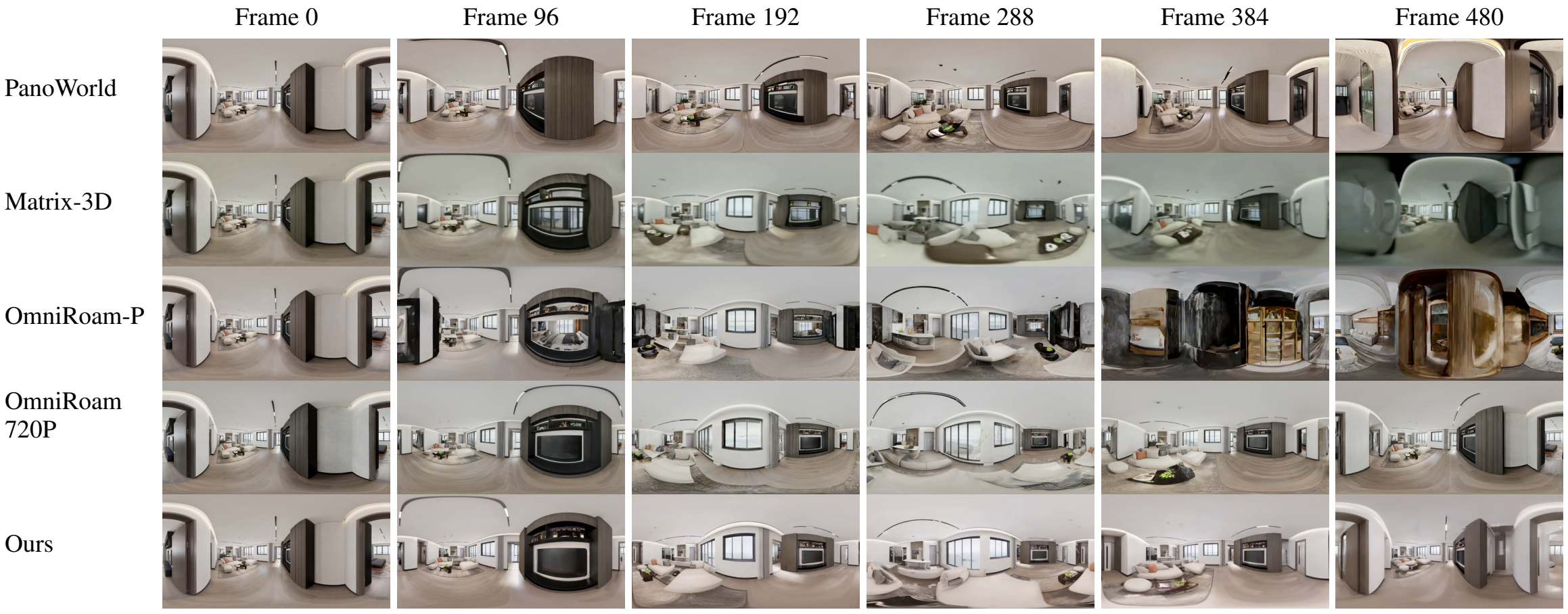

    \centering
    \begingroup
    \setlength{\tabcolsep}{1pt}
    \renewcommand{\arraystretch}{0.8}
    \newcommand{\realseeframe}[2]{\raisebox{-0.5\height}{\includegraphics[width=\dimexpr(\textwidth-54pt)/6\relax]{Fig/zeroshot_realsee3d_scene_00124/#1/#2.png}}}
    \scriptsize
    \begin{tabular}{@{}p{42pt}cccccc@{}}
        & Frame 0 & Frame 96 & Frame 192 & Frame 288 & Frame 384 & Frame 480 \\[2pt]
        PanoWorld & \realseeframe{PanoWorld}{0} & \realseeframe{PanoWorld}{96} & \realseeframe{PanoWorld}{192} & \realseeframe{PanoWorld}{288} & \realseeframe{PanoWorld}{384} & \realseeframe{PanoWorld}{480} \\[2pt]
        Matrix-3D & \realseeframe{Matrix-3D}{0} & \realseeframe{Matrix-3D}{96} & \realseeframe{Matrix-3D}{192} & \realseeframe{Matrix-3D}{288} & \realseeframe{Matrix-3D}{384} & \realseeframe{Matrix-3D}{480} \\[2pt]
        OmniRoam-P & \realseeframe{OmniRoam-P}{0} & \realseeframe{OmniRoam-P}{96} & \realseeframe{OmniRoam-P}{192} & \realseeframe{OmniRoam-P}{288} & \realseeframe{OmniRoam-P}{384} & \realseeframe{OmniRoam-P}{480} \\[2pt]
        \shortstack[l]{OmniRoam\\720P} & \realseeframe{OmniRoam}{0} & \realseeframe{OmniRoam}{96} & \realseeframe{OmniRoam}{192} & \realseeframe{OmniRoam}{288} & \realseeframe{OmniRoam}{384} & \realseeframe{OmniRoam}{480} \\[2pt]
        Ours & \realseeframe{Ours}{0} & \realseeframe{Ours}{96} & \realseeframe{Ours}{192} & \realseeframe{Ours}{288} & \realseeframe{Ours}{384} & \realseeframe{Ours}{480}
    \end{tabular}
    \endgroup
    \caption{Zero-shot qualitative comparison on RealSee3D. We show every 96th frame along the 481-frame closed-loop trajectory. Frame 480 returns to the initial viewpoint, enabling direct assessment of scene consistency upon revisitation. OmniRoam is evaluated at 720P; all other methods are evaluated at 480P.}
    \label{fig:zeroshot-realsee3d}
\end{figure}

\paragraph{Zero-Shot Evaluation.}
We further evaluate zero-shot generalization on RealSee3D under the protocol described above, with quantitative results reported in Tab.~\ref{tab:zeroshot-realsee3d}. OmniRoam-P is evaluated through autoregressive rollout. Since the full OmniRoam pipeline requires a fixed 8-segment input, we extend the 481-frame closed-loop trajectory to 641 frames by retracing the return path, generate its stride-8 sparse preview over the extended trajectory, and refine the original 481 frames for evaluation. Across both resolutions, PVWander achieves the strongest overall zero-shot performance, demonstrating robust scene consistency upon revisitation and higher trajectory fidelity on unseen scenes.

Fig.~\ref{fig:zeroshot-realsee3d} provides a qualitative comparison on one RealSee3D scene. We primarily visualize the $960\times480$ setting, which is supported by the largest set of compared methods, while additionally including the full OmniRoam result at its native $1440\times720$ resolution. Matrix-3D exhibits pronounced structural distortions, while OmniRoam-P suffers severe structural degradation under autoregressive rollout. With sparse preview guidance spanning the trajectory, the full OmniRoam pipeline better preserves the scene layout, although scene drift remains upon returning to the starting viewpoint. PanoWorld retains a plausible visual appearance, but the doorway and partition differ noticeably from those in the initial view. In comparison, PVWander better preserves the overall entrance and living-room layout upon revisitation.

\paragraph{Comparison with EvoWorld.}
EvoWorld~\citep{wangEvoWorldEvolvingPanoramic2025} also uses an evolving 3D point-cloud memory to guide panoramic video generation along camera trajectories. Its SVD backbone generates 25 frames per segment, while its Unity\_Curve test episodes are too short for multi-segment evaluation with our 81-frame Wan model. For a fair comparison, we train an SVD-based variant of PVWander and follow EvoWorld's 25-frame segment protocol. As shown in Tab.~\ref{tab:evoworld-comparison}, this variant outperforms EvoWorld across all metrics for both the first segment and the three-segment rollout. See Appendix~\ref{sec:appendix-evoworld-qualitative} for qualitative comparisons.
Furthermore, to test whether our designs can benefit an independent framework, we reproduce EvoWorld and integrate Path-Facing (PF) and Timestep-aware Gating (TAG) into it. Under identical training and inference settings at 640×320, adding PF and TAG improves all metrics over the reproduced EvoWorld baseline. After high-resolution fine-tuning at 1024×512, the enhanced model also outperforms the official EvoWorld.

\begin{table*}[ht]
    \centering
    \caption{Comparison on Unity\_curve dataset. We evaluate the official model at its native resolution and after resizing its outputs to $1024\times512$. The asterisk denotes our reproduction of EvoWorld.}
    \label{tab:evoworld-comparison}
    \setlength{\tabcolsep}{3pt}
    \resizebox{\textwidth}{!}{%
    \begin{tabular}{ll*{14}{c}}
        \toprule
        \multirow{2}{*}{Method} & \multirow{2}{*}{Resolution}
        & \multicolumn{7}{c}{First Segment (25 frames)} & \multicolumn{7}{c}{3 Segments (73 frames)} \\
        \cmidrule(lr){3-9} \cmidrule(lr){10-16}
        & & SSIM$\uparrow$ & PSNR$\uparrow$ & LPIPS$\downarrow$ & FID$\downarrow$ & $\mathrm{FID}_{\mathrm{cube}}\downarrow$ & TransErr$\downarrow$ & RotErr$\downarrow$ & SSIM$\uparrow$ & PSNR$\uparrow$ & LPIPS$\downarrow$ & FID$\downarrow$ & $\mathrm{FID}_{\mathrm{cube}}\downarrow$ & TransErr$\downarrow$ & RotErr$\downarrow$ \\
        \midrule
        EvoWorld (official, native) & $1024\times576$ & 0.7745 & 18.93 & 0.3238 & 14.65 & 19.70 & 0.768 & 10.10 & 0.7480 & 16.88 & 0.4192 & 25.53 & 23.68 & 3.467 & 43.70 \\
        EvoWorld (official, resized) & $1024\times512$ & 0.7723 & 19.06 & 0.3216 & 14.54 & 19.70 & 0.768 & 10.10 & 0.7442 & 16.96 & 0.4181 & 25.30 & 23.68 & 3.467 & 43.70 \\
        PVWander (SVD) & $1024\times512$ & \textbf{0.8199} & \textbf{22.00} & \textbf{0.1991} & \textbf{12.01} & \textbf{19.52} & \textbf{0.069} & \textbf{0.262} & \textbf{0.7891} & \textbf{20.04} & \textbf{0.2597} & \textbf{13.02} & \textbf{17.42} & \textbf{1.795} & \textbf{18.15} \\
        \midrule
        EvoWorld$^*$ & $640\times320$ & 0.7321 & 19.26 & 0.3034 & 20.68 & 27.51 & 0.575 & 8.206 & 0.6905 & 16.87 & 0.4205 & 36.09 & 34.88 & 3.540 & 45.12 \\
        EvoWorld$^*$ + PF + TAG & $640\times320$ & 0.7745 & 21.19 & 0.2156 & 14.11 & 20.77 & 0.239 & 0.338 & 0.7417 & 19.38 & 0.2715 & 16.97 & 18.81 & 1.836 & 18.27 \\
        EvoWorld$^*$ + PF + TAG & $1024\times512$ & 0.8091 & 21.13 & 0.2265 & 11.38 & 17.89 & 0.233 & 0.274 & 0.7834 & 19.43 & 0.2705 & 11.50 & 15.06 & 1.827 & 18.32 \\
        \bottomrule
    \end{tabular}}
\end{table*}

\subsection{Ablation Studies}
Unless otherwise specified, experiments in this subsection are evaluated over the full eight-segment (641-frame) rollout. Component ablations use the first-stage training setting at $640\times320$ described in Appendix~\ref{sec:appendix-experimental-settings}.

\begin{table}[h]
    \centering
    \caption{Overall ablation of PVWander. The first row denotes the ERP-adapted Wan2.1-Fun camera-only baseline. Path-Facing, when enabled, is used during both training and inference.}
    \label{tab:overall-ablation}
    \resizebox{\linewidth}{!}{
    \begin{tabular}{ccccccccc}
        \toprule
        \multicolumn{4}{c}{Components} & \multicolumn{5}{c}{8 Segments (641 frames)} \\
        \cmidrule(lr){1-4} \cmidrule(lr){5-9}
        Path-Facing & Fixed Memory & Online Memory & Timestep-aware gating & SSIM$\uparrow$ & PSNR$\uparrow$ & LPIPS$\downarrow$ & FID$\downarrow$ & $\mathrm{FID}_{\mathrm{cube}}\downarrow$ \\
        \midrule
        -- & -- & -- & -- & 0.5492 & 15.36 & 0.4220 & 24.82 & 45.21 \\
        \checkmark & -- & -- & -- & 0.5761 & 15.91 & 0.4094 & 24.72 & 42.98 \\
        -- & \checkmark & -- & -- & 0.5280 & 14.76 & 0.4460 & 29.80 & 51.39 \\
        -- & -- & \checkmark & -- & 0.5806 & 15.95 & 0.4132 & 26.69 & 42.67 \\
        -- & -- & \checkmark & \checkmark & 0.5770 & 16.06 & 0.4042 & 23.79 & 40.06 \\
        \checkmark & -- & \checkmark & -- & 0.5767 & 15.95 & 0.4107 & \textbf{23.09} & 40.50 \\
        \checkmark & -- & \checkmark & \checkmark & \textbf{0.5867} & \textbf{16.25} & \textbf{0.4032} & 23.75 & \textbf{38.13} \\
        \bottomrule
    \end{tabular}}
\end{table}

\paragraph{Overall Ablation.}
We first evaluate the main design choices of PVWander, as shown in Tab.~\ref{tab:overall-ablation}. Path-Facing improves long-horizon generation independently of 3D memory, demonstrating the benefit of canonicalizing diverse motion--orientation relationships. Timestep-aware gating further improves memory-guided generation, supporting its effectiveness in adaptively regulating imperfect reprojected guidance during denoising. Combining Path-Facing and Timestep-aware gating with online 3D memory provides the best overall balance across the evaluated metrics.
We also observe that a fixed 3D memory constructed only from the initial panorama becomes increasingly limited over extended rollout. 
This may reflect reduced reprojection coverage as the camera explores beyond the initial observation, whereas online updates incorporate newly observed regions (see Appendix~\ref{sec:appendix-fixed-online-memory}).

\paragraph{Timestep-aware gating Modulation.}
\begin{table}[ht]
    \centering
    \begin{minipage}[t]{0.62\linewidth}
    \vspace{0pt}
    \centering
    \caption{Param-only uses timestep-independent learnable modulation; Scalar-wise uses a shared timestep-dependent scalar, while Channel-wise predicts a gate per channel.}
    \label{tab:gating-ablation}
    \footnotesize
    \setlength{\tabcolsep}{1.5pt}
    \renewcommand{\arraystretch}{1.5}
    \begin{tabular*}{\linewidth}{@{\extracolsep{\fill}}clccccc@{}}
        \toprule
        Memory & Gating & SSIM$\uparrow$ & PSNR$\uparrow$ & LPIPS$\downarrow$ & FID$\downarrow$ & FID$_{\mathrm{cube}}\downarrow$ \\
        \midrule
        -- & -- & 0.5492 & 15.36 & 0.4220 & 24.82 & 45.21 \\
        \checkmark & None & \textbf{0.5806} & 15.95 & 0.4132 & 26.69 & 42.67 \\
        \checkmark & Param-only & 0.5605 & 15.81 & 0.4126 & 26.18 & 47.53 \\
        \checkmark & Scalar-wise & 0.5750 & 15.99 & 0.4092 & 26.61 & 46.26 \\
        \checkmark & Channel-wise & 0.5770 & \textbf{16.06} & \textbf{0.4042} & \textbf{23.79} & \textbf{40.06} \\
        \bottomrule
    \end{tabular*}
    \end{minipage}\hfill
    \begin{minipage}[t]{0.36\linewidth}
    \vspace{6pt}
    \centering
    \includegraphics[width=\linewidth,height=0.64\linewidth,keepaspectratio]{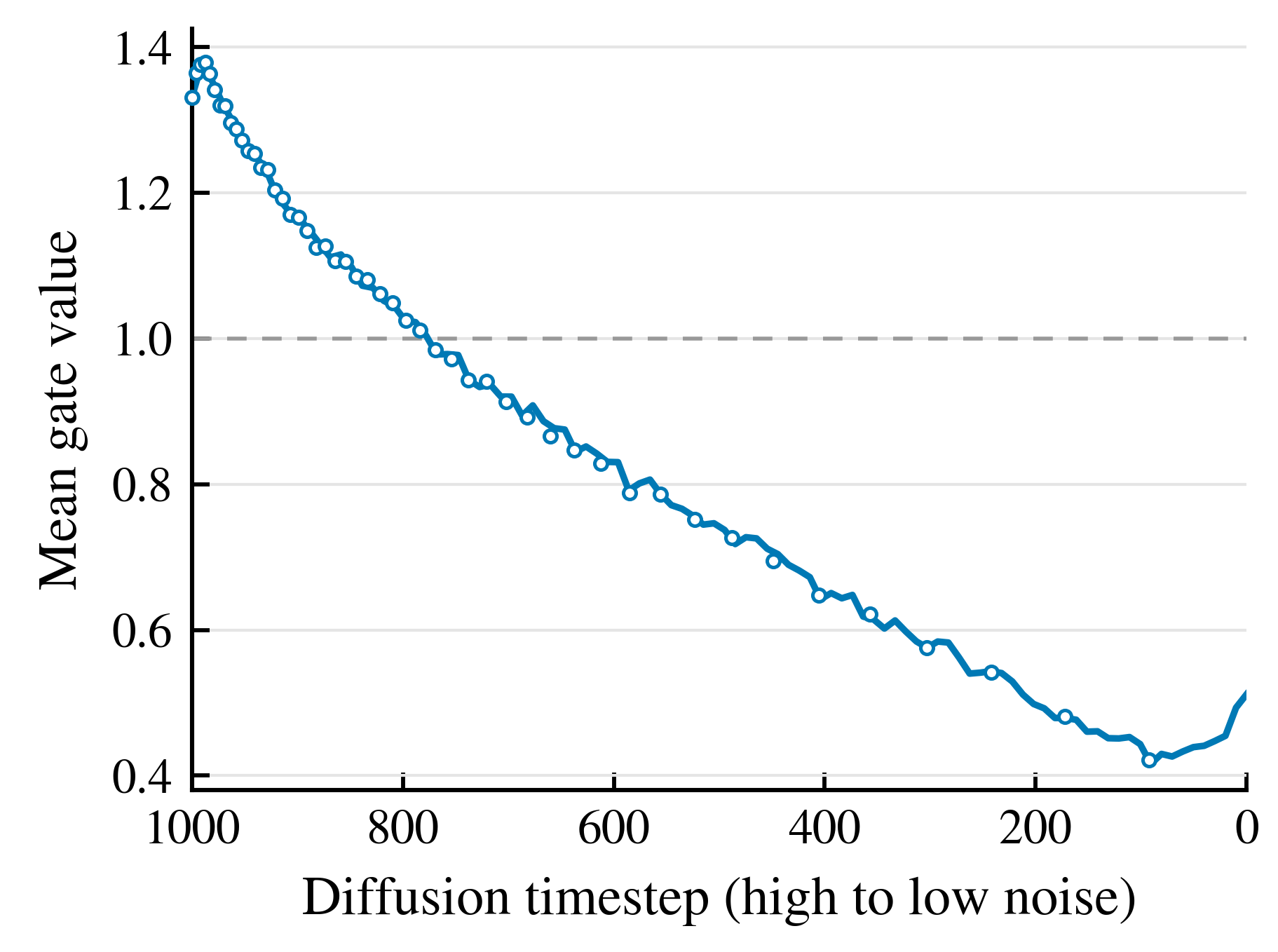}
    \captionof{figure}{Learned gate values averaged across channels. Open circles mark the 50-step inference schedule with a shift of 5.}
    \label{fig:gating-timestep}
    \end{minipage}
\end{table}

We examine whether the benefit of memory modulation arises from timestep awareness rather than simply introducing additional learnable parameters. As shown in Tab.~\ref{tab:gating-ablation}, the proposed Timestep-aware gating achieves the strongest overall performance among the gating variants. The parameter-only variant provides no consistent improvement over ungated memory, indicating that the gain is not merely due to the additional modulation parameters. Using a timestep-dependent scalar provides only limited improvement, whereas the proposed channel-wise formulation allows the modulation to adapt more flexibly across denoising stages. We further compare against manually specified render-conditioning cutoffs in Appendix~\ref{sec:appendix-hard-cutoff}, where the learned gating also outperforms the tested hard-cutoff configurations.

Fig.~\ref{fig:gating-timestep} visualizes the learned modulation across denoising timesteps. The mean gate is stronger during the early high-noise stage and generally decreases during later refinement. This learned trend is consistent with our motivation to emphasize memory-based structural guidance early while reducing reliance on imperfect rendered details at later stages.

\paragraph{Zero-Shot Transfer of Path-Facing Canonicalization.}
Since Path-Facing canonicalization introduces no additional parameters or architectural changes, it can be applied to existing panoramic generators without additional fine-tuning. Here we evaluate its zero-shot transfer on InteriorGS-OmniRoam using Wan2.1-Fun, OmniRoam, Matrix-3D, and PanoWorld.

As shown in Tab.~\ref{tab:path-facing-zeroshot}, Path-Facing improves SSIM in all five configurations, with gains in PSNR and LPIPS in most cases.
These results show that the benefits of Path-Facing are not limited to models trained with this canonicalization.
Notably, both Wan2.1-Fun and OmniRoam have been fine-tuned on InteriorGS-OmniRoam with fixed ERP orientations. Their improvements support the benefit of trajectory-dependent alignment beyond fixed orientations alone. For Wan2.1-Fun, the gains are larger when Path-Facing is used during both training and inference (Tab.~\ref{tab:overall-ablation}).

\begin{table}[ht]
    \centering
    \caption{Zero-shot transfer of Path-Facing on InteriorGS-OmniRoam. Each entry reports performance before $\rightarrow$ after applying Path-Facing at inference. All evaluated checkpoints are trained without Path-Facing. OmniRoam (Preview) denotes the stride-8 sparse preview; all others use the full 641-frame rollout.}
    \label{tab:path-facing-zeroshot}
    \small
    \setlength{\tabcolsep}{6pt}
    \begin{tabular}{@{}lccc@{}}
        \toprule
        Method & SSIM$\uparrow$ & PSNR$\uparrow$ & LPIPS$\downarrow$ \\
        \midrule
        Wan2.1-Fun
        & 0.5492 $\rightarrow$ \textbf{0.5600}
        & 15.36 $\rightarrow$ \textbf{15.64}
        & 0.4220 $\rightarrow$ \textbf{0.4201} \\

        OmniRoam
        & 0.6063 $\rightarrow$ \textbf{0.6092}
        & \textbf{14.59} $\rightarrow$ 14.58
        & 0.4662 $\rightarrow$ \textbf{0.4608} \\

        OmniRoam (Preview)
        & 0.5823 $\rightarrow$ \textbf{0.5854}
        & 15.02 $\rightarrow$ \textbf{15.04}
        & 0.4392 $\rightarrow$ \textbf{0.4330} \\

        Matrix-3D
        & 0.5748 $\rightarrow$ \textbf{0.5776}
        & 12.76 $\rightarrow$ \textbf{12.95}
        & \textbf{0.5765} $\rightarrow$ 0.5770 \\

        PanoWorld
        & 0.4546 $\rightarrow$ \textbf{0.4650}
        & 11.96 $\rightarrow$ \textbf{12.21}
        & 0.5913 $\rightarrow$ \textbf{0.5903} \\
        \bottomrule
    \end{tabular}
\end{table}

\begin{wrapfigure}{r}{0.45\textwidth}
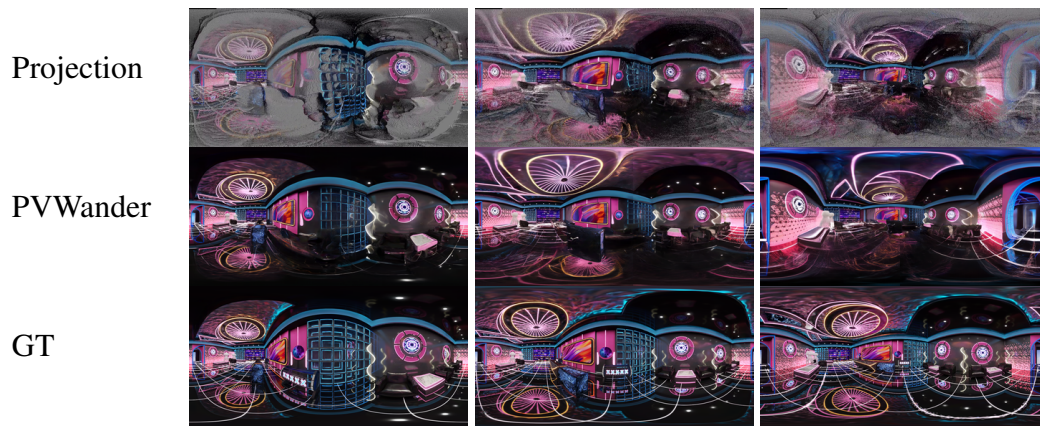

    \vspace{-10pt}
    \centering
    \begingroup
    \setlength{\tabcolsep}{0.7pt}
    \renewcommand{\arraystretch}{0.7}
    \newcommand{\limframe}[2]{\raisebox{-0.5\height}{\includegraphics[width=\dimexpr(\linewidth-38pt)/3\relax]{Fig/limitation_vis/#1/#2.png}}}
    \tiny
    \begin{tabular}{@{}p{32pt}ccc@{}}
        Projection & \limframe{DepthProjection}{81} & \limframe{DepthProjection}{201} & \limframe{DepthProjection}{321} \\[1pt]
        PVWander & \limframe{PVWander}{81} & \limframe{PVWander}{201} & \limframe{PVWander}{321} \\[1pt]
        GT & \limframe{GT}{81} & \limframe{GT}{201} & \limframe{GT}{321}
    \end{tabular}
    \endgroup
    \caption{Failure case with severely corrupted 3D-memory geometry. PVWander preserves plausible appearance but shows structural deviations.}
    \label{fig:limitations}
\end{wrapfigure}

\paragraph{Limitations.}
PVWander relies on estimated depth to construct and update its 3D memory. 
Depth estimation errors, particularly in geometrically ambiguous regions such as reflective surfaces, can lead to severely corrupted memory reprojections.
Although train--inference consistency and Timestep-aware gating make the model more robust to imperfect projected guidance, they cannot fully compensate when depth errors substantially alter the underlying scene structure.
As shown in Fig.~\ref{fig:limitations}, the model recovers plausible appearance from distorted memory projections, but structural deviations remain. 
Future improvements in geometric estimation could further strengthen structural fidelity.

\section{Conclusion}
This paper introduced PVWander, a framework for long-horizon panoramic video wandering that continuously accumulates generated observations into an online 3D memory and reuses the accumulated scene information to guide subsequent generation.
Path-Facing canonicalization simplifies diverse motion--orientation relationships without introducing additional parameters and with negligible computational overhead, while Timestep-aware gating modulation adaptively regulates imperfect memory guidance across denoising stages.
Experiments on InteriorGS-OmniRoam and zero-shot evaluation on RealSee3D demonstrate improved long-horizon generation quality, scene consistency, and trajectory fidelity.
Ablation studies validate the effectiveness of the proposed designs, while transfer experiments further show that Path-Facing canonicalization can benefit existing panoramic generators without additional training.
While the current framework focuses on static-scene exploration, extending memory-guided panoramic generation to dynamic environments with object motion and interactions remains an important direction toward more general world models.

\bibliography{ref_final}
\bibliographystyle{iclr2027_conference}

\clearpage
\appendix
\section{Appendix}

\subsection{Preliminaries}
\label{sec:appendix-preliminaries}
\paragraph{Video Diffusion Model.}
We adopt Wan2.1~\citep{wanWanOpenAdvanced2025} as our video generation backbone. Given a video $\mathbf{x}$, a spatiotemporal VAE encodes it into a latent representation $\mathbf{z}_{\mathrm{clean}}=\mathcal{E}(\mathbf{x})$.
Following flow matching~\citep{lipmanFlowMatchingGenerative2023}, we construct noisy latents from Gaussian noise $\boldsymbol{\epsilon}\sim\mathcal{N}(\mathbf{0},\mathbf{I})$ and the clean latent as $\mathbf{z}_{\sigma}=(1-\sigma)\mathbf{z}_{\mathrm{clean}}+\sigma\boldsymbol{\epsilon}$, where $\sigma\in[0,1]$.
The model uses the scheduler timestep $t=1000\sigma$ and predicts the target velocity $\mathbf{v}=\boldsymbol{\epsilon}-\mathbf{z}_{\mathrm{clean}}$ with
\begin{equation}
    \mathcal{L}
    =
    \mathbb{E}
    \left[
        \left\|
        \mathbf{u}_{\theta}(\mathbf{z}_{\sigma},\mathbf{c},t)
        -
        \mathbf{v}
        \right\|_2^2
    \right],
\end{equation}
where $\mathbf{c}$ denotes the generation conditions.

To enable camera-controllable video generation, our model is initialized from \textit{Wan2.1-Fun-V1.1-1.3B-Control-Camera}. Given a sequence of camera poses, the corresponding pixel-wise Plücker ray maps are used as camera conditions. These ray maps are encoded by a lightweight camera adapter and additively fused with the patch-embedded latent features, providing explicit conditioning for camera-motion control.

\paragraph{3D Memory-Guided Video Generation.}
3D-guided video generation provides explicit spatial context by lifting observed images into 3D representation, and reprojecting them to future viewpoints as additional conditions for video generation~\citep{yuViewCrafterTamingVideo2025,renGEN3C3DInformedWorldConsistent2025}. 
For long-horizon generation, some recent methods~\citep{wuVideoWorldModels2025} further maintain an online 3D memory to accumulate newly generated observations, which can then be reprojected and reused as conditioning for subsequent videos.

\subsection{ERP-aware Camera Control}
\label{sec:appendix-erp-camera}
We build our camera-controllable panoramic video generation baseline on \textit{Wan2.1-Fun-V1.1-1.3B-Control-Camera}. Although this model provides an effective camera-conditioning interface, its original Pl\"ucker representation is constructed from perspective camera rays, which cannot be directly applied to ERP panoramas. We therefore adapt its camera representation to ERP videos by constructing pixel-wise spherical rays, similar to CamPVG~\citep{jiCamPVGCameraControlledPanoramic2025}.

For ERP panoramas, pixel-wise ray directions are determined by spherical coordinates in the local camera frame.
Given a pixel $(u,v)$ in an ERP frame of resolution $H\times W$, we define its longitude and latitude as
$\lambda=2\pi(u/W-1/2)$ and
$\varphi=\pi(1/2-v/H)$, respectively. The corresponding unit spherical ray direction is
\begin{equation}
\mathbf{d}(u,v)
=
\begin{bmatrix}
\cos\varphi\sin\lambda,\;
\sin\varphi,\;
\cos\varphi\cos\lambda
\end{bmatrix}^{\top}.
\end{equation}
Let $\mathbf{c}_t$ denote the camera position at frame $t$. We define its relative position with respect to the first frame as
$\mathbf{o}_t=\mathbf{c}_t-\mathbf{c}_0$,
and construct the panoramic Pl\"ucker embedding as
\begin{equation}
\mathbf{p}_t(u,v)
=
\left[
\mathbf{o}_t \times \mathbf{d}(u,v),\;
\mathbf{d}(u,v)
\right].
\end{equation}
This formulation retains the pixel-wise Pl\"ucker conditioning interface of the pretrained camera adapter while incorporating the spherical projection geometry of ERP panoramas.

\subsection{Panoramic Representation and ERP Reorientation}
\label{sec:appendix-panorama-conversion}

\paragraph{Cubemap--ERP representation.}

A panoramic observation captures the complete $360^\circ$ surrounding view of a camera. Such spherical content can be represented using different projection formats. Cubemap and equirectangular projection (ERP) are two commonly used parameterizations of the viewing sphere.

Cubemap represents the spherical view using multiple perspective images, where each view corresponds to a limited field of view. In contrast, ERP unwraps the entire sphere into a single rectangular image by mapping longitude and latitude to the horizontal and vertical image axes, respectively. Therefore, the full panoramic content, including front, back, left, right, up, and down directions, is stored in one continuous image.

\begin{figure}[ht]
    \centering
    \includegraphics[width=0.9\linewidth]{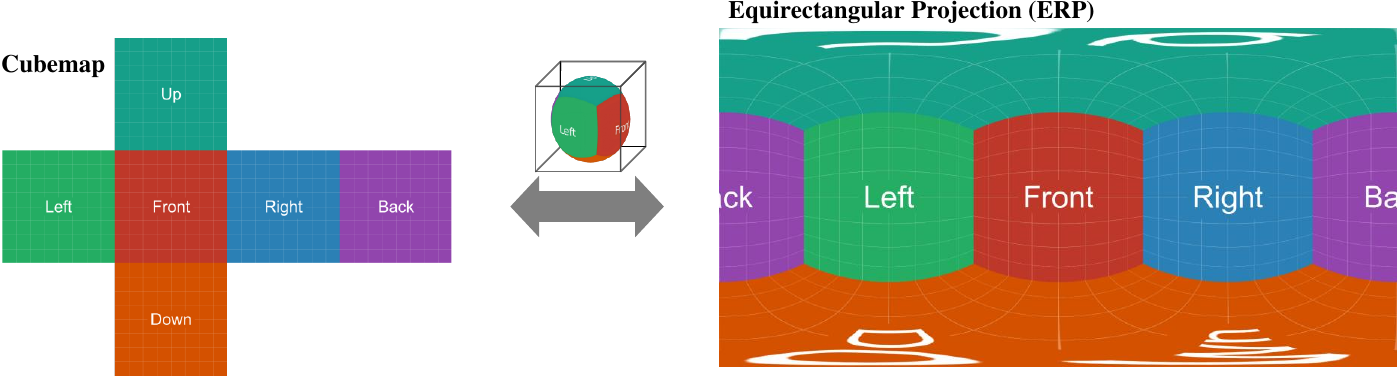}
    \caption{Cubemap and ERP represent the same viewing sphere with different sampling grids. Colors identify the six Cubemap faces and their corresponding regions in ERP.}
    \label{fig:cubemap-to-erp}
\end{figure}

Due to this property, ERP provides a convenient representation for panoramic video generation models, as temporal modeling and image-based generative models can directly operate on a single image sequence. Fig.~\ref{fig:cubemap-to-erp} illustrates the correspondence between Cubemap and ERP representations.

\paragraph{ERP reorientation.}

Because ERP covers the complete viewing sphere, changing its horizontal reference orientation does not alter the underlying scene content. A yaw rotation around the vertical axis only shifts the longitude coordinate while preserving latitude, which corresponds to a horizontal circular shift in ERP space.

For an ERP image \(I\), a yaw offset \(\alpha\) can be applied as

\[
I_{\alpha}(u,v)
=
I\left(\left(u-\frac{W\alpha}{2\pi}\right)\bmod W,v\right),
\]

where \(W\) denotes the ERP width and \(\alpha\) is the yaw offset in radians.

Therefore, yaw reorientation preserves spherical coverage and only changes the horizontal reference orientation of the panorama. This cyclic property enables \textbf{Path-Facing} to canonicalize each clip's ERP reference orientation according to its principal translation direction, without modifying the prescribed camera trajectory.

\begin{figure}[ht]
    \centering
    \includegraphics[width=0.95\linewidth]{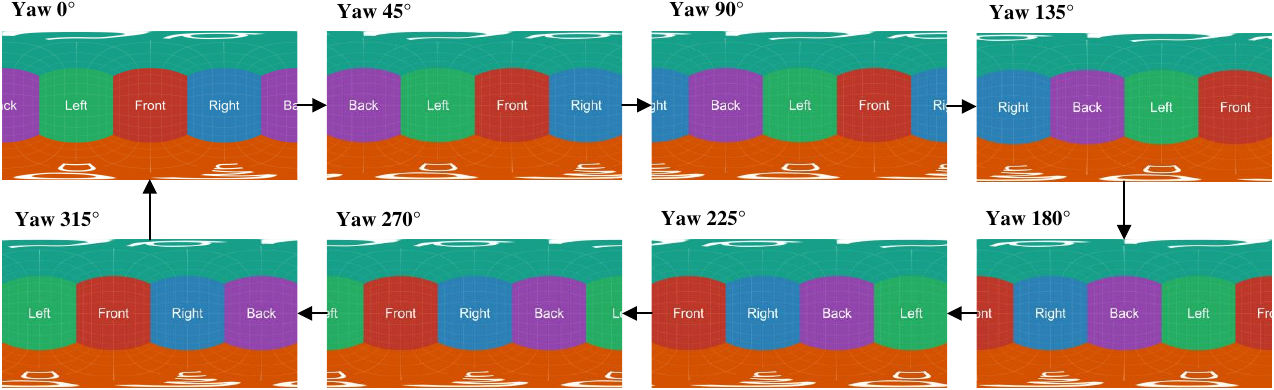}
    \caption{ERP panoramas under different yaw reorientations. Horizontal circular shifts only change the horizontal reference orientation while preserving the complete spherical content.}
    \label{fig:erp-rotation}
\end{figure}

\subsection{Experimental Settings and Implementation Details}
\label{sec:appendix-experimental-settings}

\paragraph{Datasets and Evaluation Protocols.}
We conduct our main experiments on the \textit{InteriorGS-OmniRoam} panoramic video dataset~\citep{liuOmniRoamWorldWandering2026}, which is rendered from around 1,000 3D Gaussian Splatting scenes collected from InteriorGS~\citep{miaoPhysicallyExecutable3D2025}. 
 We follow the train--test split of OmniRoam, using 876 scenes for training and 98 scenes for evaluation. Each scene contains a sequence of 800 panoramic frames rendered along a wandering camera trajectory. The dataset uses a fixed ERP orientation across its diverse camera trajectories, taking advantage of the full $360^\circ$ coverage of panoramic observations.
For evaluation, following OmniRoam's protocol, we recursively generate 8 consecutive segments, resulting in a 641-frame sequence. 
We report results for both the first segment and the full rollout. For generation with Path-Facing canonicalization, we restore the generated panoramas to the GT orientations before evaluation to ensure a fair comparison with other methods.
Appearance fidelity is evaluated using PSNR, SSIM, and LPIPS, while perceptual quality is measured using FID. Since directly resizing panoramic frames to square images distorts ERP geometry, we additionally report $\mathrm{FID}_{\mathrm{cube}}$ following SMGD~\citep{sunSphericalManifoldGuided2025} and PanoWorld~\citep{liPanoWorldRealworldPanoramic2026}, by projecting each panorama onto six cubemap faces, computing FID for each face, and averaging the results. 
For camera controllability, following~\citep{liPanoWorldRealworldPanoramic2026,duanWorldScoreUnifiedEvaluation2025}, we further evaluate trajectory fidelity by using ViPE~\citep{huangViPEVideoPose2025} to reconstruct frame-wise camera poses from the generated videos and compare the recovered trajectories with the prescribed camera trajectories. We report rotation error (\textit{RotErr}, degrees) and scale-aligned translation error (\textit{TransErr}). Appendix~\ref{sec:appendix-camera-metrics} provides the detailed pose alignment and metric definitions.

To further evaluate zero-shot generalization, we conduct experiments on RealSee3D~\citep{liArgusMetricPanoramic2026}. Since RealSee3D consists of static RGB-D captures rather than video sequences, following~\citep{suGenieSimPanoWorld2026}, we use initial panoramas from 90 scenes and use the GT depth of each initial panorama to construct a 481-frame camera trajectory that returns to the starting viewpoint at the final frame. This enables loop-closure evaluation without requiring GT roaming videos. We therefore report \textit{SSIM-Loop}, \textit{PSNR-Loop}, and \textit{LPIPS-Loop} between the initial and final frames to evaluate loop-closure consistency, while \textit{TransErr} and \textit{RotErr} are computed over the full recovered camera trajectory to evaluate trajectory fidelity. All models are evaluated on RealSee3D without any fine-tuning.

\paragraph{Implementation Details.}
We initialize PVWander from \textit{Wan2.1-Fun-V1.1-1.3B-Control-Camera} and train it on four NVIDIA A100 GPUs using 81-frame clips. We use rank-64 LoRA for the attention layers and jointly train the conditioning modules and input projections. We use AdamW with BF16 mixed precision and an effective batch size of 32.
Training is conducted in three stages with progressively increasing ERP resolution. We first train at $640\times320$ for 5,000 steps with a learning rate of $10^{-4}$, using camera conditioning only for the first 1,000 steps and introducing memory-render guidance for the remaining 4,000 steps. We then train at $960\times480$ for 3,500 steps with a learning rate of $10^{-5}$, where the temporal stride is fixed to 1 for the first 1,500 steps and uniformly sampled from $\{1,2,3\}$ for each clip during the remaining 2,000 steps to improve robustness to varying temporal dynamics. Finally, we train at $1440\times720$ for 500 steps with a learning rate of $5\times10^{-6}$ while retaining random temporal-stride sampling.
We use PVMetric to estimate depth for 3D memory construction during both training and inference, maintaining train--inference consistency and allowing the generator to adapt to its prediction errors during training. Appendix~\ref{sec:appendix-depth-consistency} further demonstrates the importance of this consistency for generation quality.
For Path-Facing, we choose the sign of the first principal component so that projected camera positions have positive correlation with frame indices. When this correlation is numerically negligible, we resolve the sign using the displacement from the first to the last frame.
During training, we apply Path-Facing to 90\% of clips. For the remaining 10\%, we use the horizontal direction from the first camera position to a randomly selected target frame's position as the common ERP reference orientation for all frames to improve robustness to orientation variation.
For memory reprojection, we randomly sample the number of source frames from 1 to 4 during training. At inference, we use four source frames for each segment except the first.

\paragraph{Training for the EvoWorld comparison.}
On Unity\_Curve, we train three models at $640\times320$ for 5,000 steps with a batch size of 24: SVD-based PVWander, our EvoWorld reproduction, and its PF+TAG variant. We then fine-tune SVD-based PVWander and EvoWorld with PF+TAG at $1024\times512$ for another 3,000 steps (Tab.~\ref{tab:evoworld-comparison}). Other settings follow the official EvoWorld implementation.

\subsection{Qualitative Comparison with EvoWorld}
\label{sec:appendix-evoworld-qualitative}
\begin{figure}[!htb]
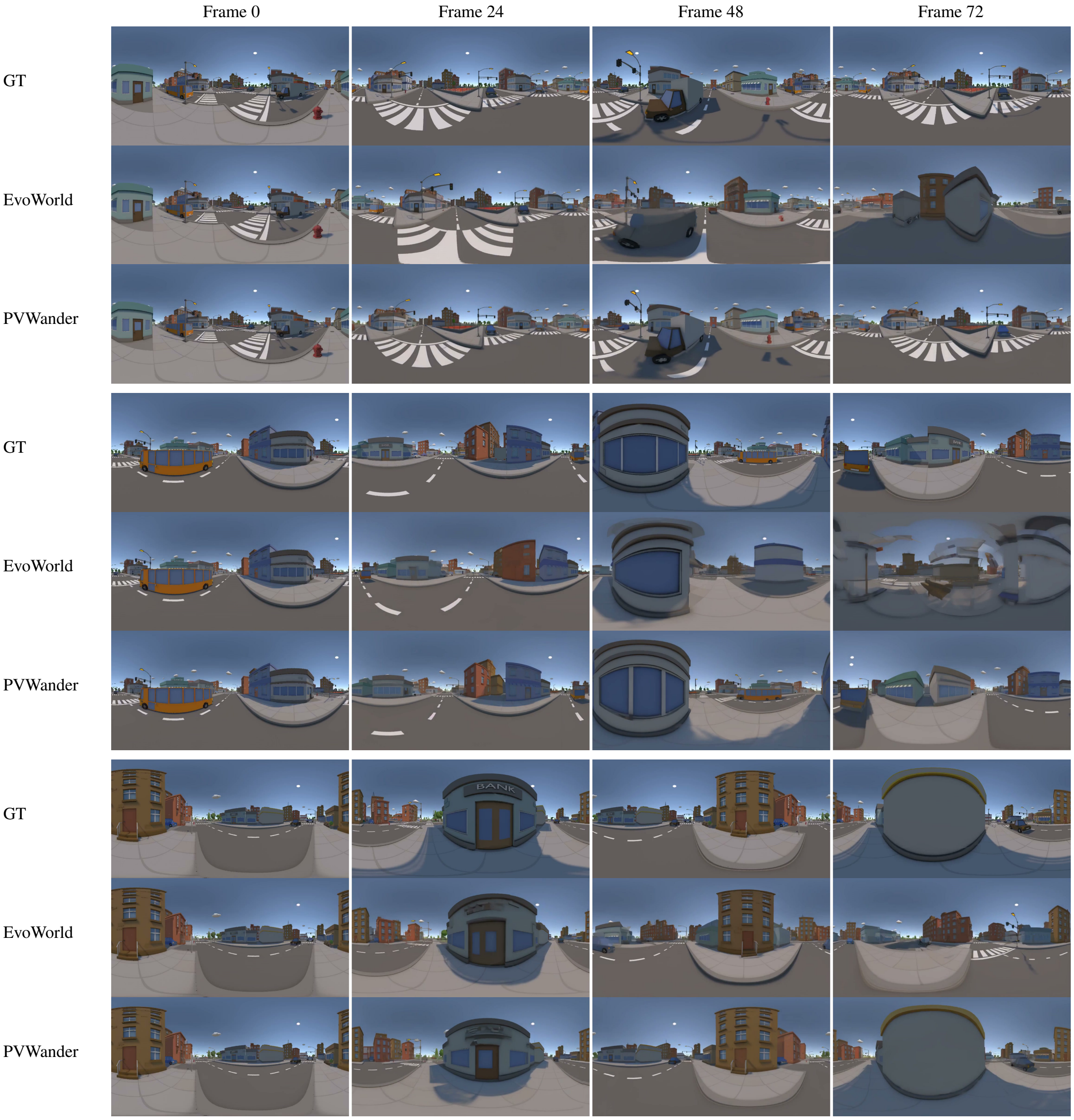

    \centering
    \scriptsize
    \setlength{\tabcolsep}{1pt}
    \renewcommand{\arraystretch}{0.8}
    \newcommand{\evowidth}{\dimexpr\linewidth-44pt\relax}
    \newcommand{\evostrip}[2]{\raisebox{-0.5\height}{\includegraphics[width=\evowidth]{Fig/evoworld_unity_curve/#1_#2.jpg}}}
    \newcommand{\evoepisode}[1]{%
        GT & \evostrip{#1}{gt} \\[1pt]
        EvoWorld & \evostrip{#1}{evo} \\[1pt]
        PVWander & \evostrip{#1}{pv} \\}
    \begin{tabular}{@{}p{42pt}c@{}}
        & \makebox[\evowidth]{\makebox[\dimexpr(\textwidth-44pt)/4\relax]{Frame 0}\makebox[\dimexpr(\textwidth-44pt)/4\relax]{Frame 24}\makebox[\dimexpr(\textwidth-44pt)/4\relax]{Frame 48}\makebox[\dimexpr(\textwidth-44pt)/4\relax]{Frame 72}} \\[2pt]
        \evoepisode{0641}
        \addlinespace[4pt]
        \evoepisode{0007}
        \addlinespace[4pt]
        \evoepisode{0238}
    \end{tabular}
    \caption{Qualitative comparison on three Unity\_Curve test trajectories. Each group shows ground truth, official EvoWorld, and SVD-based PVWander at the initial frame and the end of each of the three segments.}
    \label{fig:evoworld-qualitative}
\end{figure}

Figure~\ref{fig:evoworld-qualitative} shows the official EvoWorld model and SVD-based PVWander alongside the ground truth for three Unity\_Curve trajectories. In the bottom example, the rounded facade occupies the center of the final ground-truth panorama. PVWander reproduces its position and scale more closely, while EvoWorld shows buildings farther down the street.

\subsection{Path-Facing on Challenging Trajectories}
\label{sec:appendix-pf-challenging}
Path-Facing uses PCA on the horizontal camera positions over the entire segment to choose a shared ERP orientation. If no reliable principal direction can be determined, it uses the orientation of the first frame. For pure rotation, the camera center is fixed, so the requested views can be obtained directly by spherical reprojection of the initial panorama.

In this section, we further evaluate Path-Facing on trajectories where the camera changes direction within a segment or the dominant translation direction is ambiguous. Specifically, we render ground-truth panoramas from the 3DGS representations of the 98 InteriorGS-OmniRoam test scenes along 240 trajectories: 80 U-turns, 80 closed-return trajectories with near-zero net displacement, and 80 zig-zag trajectories. Each trajectory contains 81 frames. 

To isolate Path-Facing, we compare Wan2.1-Fun with two variants that apply Path-Facing only at inference (zero-shot) or during both training and inference.

\begin{table}[htbp]
    \centering
    \caption{Evaluation of Path-Facing under challenging trajectories.}
    \label{tab:pf-challenging}
    \begin{tabular}{lcccc}
        \toprule
        Wan2.1-Fun configuration & SSIM$\uparrow$ & PSNR$\uparrow$ & LPIPS$\downarrow$ & FID$\downarrow$ \\
        \midrule
        Baseline & 0.7725 & 23.84 & 0.0876 & 9.58 \\
        + Path-Facing at inference & 0.7736 & 23.88 & \textbf{0.0872} & 9.21 \\
        + Path-Facing at training and inference & \textbf{0.7760} & \textbf{23.95} & 0.0875 & \textbf{9.03} \\
        \bottomrule
    \end{tabular}
\end{table}

As shown in Tab.~\ref{tab:pf-challenging}, despite frequent changes in local motion direction and ambiguity in the dominant translation direction, Path-Facing maintains generation performance comparable to the baseline across these trajectories. While Fig.~\ref{fig:intro-teaser} shows that Path-Facing aligns motion directions with ERP orientations across most of the 31K sampled clips, Tab.~\ref{tab:pf-challenging} further shows that it remains applicable even when the input trajectories challenge the assumption of a dominant motion direction.

\subsection{Computational Cost and Memory Growth}
\label{sec:appendix-computational-cost}
PVWander retains observations from earlier segments, so generating each new segment requires processing a growing history. In this section, we examine how this affects per-segment runtime, then compare the total cost of an eight-segment rollout across methods.

Storing $N$ historical RGB observations of $P$ pixels each requires $O(NP)$ space. Before generating a new segment, PVWander re-estimates depth from the observation history with PVMetric to improve depth consistency, then selects at most $K=4$ source observations based on their coverage of the target views. The cost of depth estimation and source selection grows with the history length. Reprojecting the selected observations into $T$ target frames costs $O(TKP)$.

\paragraph{Stage-wise runtime.}
Table~\ref{tab:stage-runtime} reports per-segment runtimes at $1440\times720$ on an A100 40\,GB GPU. From segment 2 to segment 8, the history grows from 81 to 561 frames and depth estimation time increases from 7.53 to 50.17 seconds. Depth estimation accounts for most of the increase in memory-conditioning time. With at most four source observations, reprojection takes about 0.8 seconds at all three history lengths.

\begin{table}[htbp]
    \centering
    \caption{PVWander runtime by stage for three target segments. Times are in seconds per segment.}
    \label{tab:stage-runtime}
    \small
    \setlength{\tabcolsep}{5pt}
    \begin{tabular}{@{}rrrrrrr@{}}
        \toprule
        \shortstack[r]{Target\\segment} & \shortstack[r]{Historical\\frames} & \shortstack[r]{Depth\\estimation} & \shortstack[r]{Source\\selection} & Reprojection & \shortstack[r]{VAE encode/\\decode} & Denoising \\
        \midrule
        2 & 81  &  7.53 & 0.12 & 0.77 & 47.35 & 472.80 \\
        4 & 241 & 21.33 & 0.31 & 0.83 & 47.35 & 472.80 \\
        8 & 561 & 50.17 & 0.69 & 0.78 & 47.35 & 472.80 \\
        \bottomrule
    \end{tabular}
\end{table}

\paragraph{Full rollout cost.}
Table~\ref{tab:rollout-cost} compares the compute and runtime of eight-segment rollouts. PVWander takes 72.76 minutes, compared with 66.84 minutes for Wan2.1-Fun on the same backbone. For an eight-segment PVWander rollout, diffusion denoising, VAE encoding/decoding, and accumulated depth processing take approximately 63.04, 6.31, and 3.40 minutes, respectively. Thus, although memory-related costs grow with history length, they do not constitute the primary computational bottleneck in the current 8-segment setting.

\begin{table}[htbp]
    \centering
    \caption{Computational cost per eight-segment rollout at $1440\times720$ on an A100 40\,GB GPU.}
    \label{tab:rollout-cost}
    \small
    \begin{tabular}{@{}llrr@{}}
        \toprule
        Method & Backbone & Compute ($10^3$ TFLOPs) & Runtime (min) \\
        \midrule
        Wan2.1-Fun                   & Wan2.1-1.3B &   646 &  66.84 \\
        PVWander                     & Wan2.1-1.3B &   657 &  72.76 \\
        OmniRoam (Preview + Refine)  & Wan2.1-1.3B & 5,231 & 488.63 \\
        Matrix-3D                    & Wan2.2-5B   &   400 &  48.83 \\
        PanoWorld                    & Wan2.2-5B   &   217 &  41.54 \\
        \bottomrule
    \end{tabular}
\end{table}

Interestingly, although Matrix-3D and PanoWorld use larger 5B backbones, their computational costs and runtimes are lower. This is mainly because Wan2.2-5B employs a VAE with a higher spatial compression ratio, resulting in fewer latent tokens. OmniRoam, in contrast, concatenates Preview conditioning latents with noisy video latents, doubling the token count and substantially increasing attention computation.

\subsection{Training--Inference Depth Consistency}
\label{sec:appendix-depth-consistency}

\begin{table}[h]
    \centering
    \caption{
    Depth-source ablation for geometric conditioning.
    All results are evaluated on the first 81-frame segment.
    ``w/ align'' denotes aligning the predicted depth to the GT depth scale.
    The GT/GT setting serves as an oracle reference.
    }
    \label{tab:depth-consistency}
    \resizebox{\linewidth}{!}{
    \begin{tabular}{lllccc}
        \toprule
        Method & Train Depth & Inference Depth & SSIM$\uparrow$ & PSNR$\uparrow$ & LPIPS$\downarrow$ \\
        \midrule
        \multirow{3}{*}{Pose + 3D Memory}
        & GT       & GT (Oracle)              & 0.7716 & 22.68 & 0.1424 \\
        & GT       & PVMetric (w/ align)      & 0.7170 & 20.83 & 0.1761 \\
        & PVMetric & PVMetric                 & \textbf{0.7357} & \textbf{21.41} & \textbf{0.1654} \\
        \midrule
        \multirow{2}{*}{Matrix-3D}
        & GT       & MoGe (w/ align)           & 0.7087 & 20.48 & 0.1838 \\
        & PVMetric & PVMetric                 & \textbf{0.7301} & \textbf{21.29} & \textbf{0.1686} \\
        \bottomrule
    \end{tabular}}
\end{table}

We examine how the depth source used for geometric conditioning affects generation performance. For the Pose + 3D Memory setting, using GT depth during both training and inference provides an oracle upper bound. However, when a model trained with GT depth is evaluated using PVMetric predictions, performance drops substantially even with GT scale alignment. Using PVMetric during both training and inference improves SSIM from 0.7170 to 0.7357, PSNR from 20.83 to 21.41, and LPIPS from 0.1761 to 0.1654. This suggests that exposing the generator to prediction errors during training is beneficial when the same estimated depth is required at inference. Based on these results, PVWander uses PVMetric for 3D memory construction during both training and inference.

We also observe a similar trend for the adapted Matrix-3D setting: replacing its GT-training/MoGe-inference configuration with PVMetric for both stages improves all three metrics. We do not additionally train Matrix-3D with MoGe-predicted depth because applying MoGe to the panoramic training set is prohibitively slow. 

\begin{figure}[htbp]
    \centering
    \includegraphics[width=0.95\linewidth]{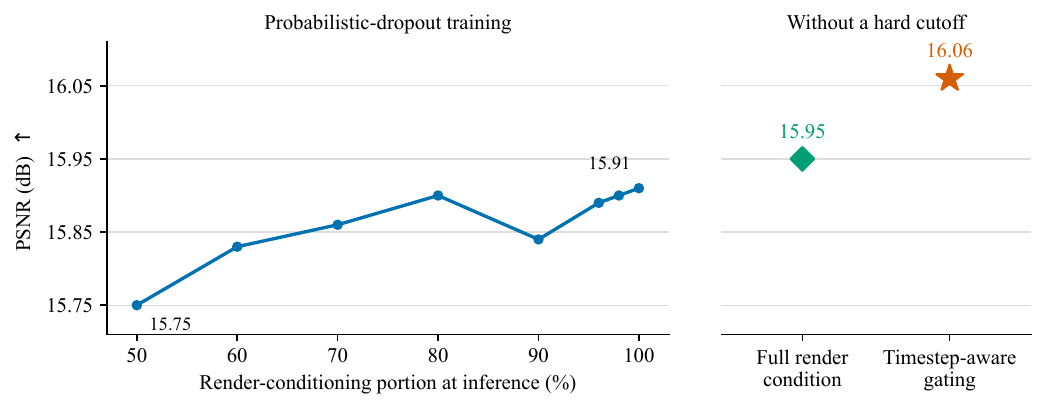}
    \caption{Hard render-conditioning cutoffs versus Timestep-aware gating on eight-segment rollouts. The curve reports PSNR for a model trained with probabilistic render-condition dropout, evaluated with different render-conditioning portions during denoising. All eight tested settings (50, 60, 70, 80, 90, 96, 98, and 100\%) are shown. The diamond and star denote full render conditioning and Timestep-aware gating, respectively; they share the PSNR axis but are displayed separately because neither uses a hard cutoff. Higher PSNR is better.}
    \label{fig:timestep-cutoff-psnr}
\end{figure}

\subsection{Hard Render-Conditioning Cutoffs vs. Timestep-aware gating}
\label{sec:appendix-hard-cutoff}
We compare manually disabling the render condition during denoising with full render conditioning and Timestep-aware gating on eight-segment rollouts. 
For the hard-cutoff baseline, training uses timestep-dependent probabilistic dropout: the render condition is disabled for timesteps below 50\% of the timestep range and retained with probability 0.5 for timesteps above 50\%. At inference, we vary the portion of the denoising process that uses the render condition from 50\% to 100\%, retaining it in the initial high-noise portion and disabling it thereafter. 
Full render conditioning uses the condition throughout both training and inference, whereas Timestep-aware gating learns channel-wise modulation without a hard cutoff.

Fig.~\ref{fig:timestep-cutoff-psnr} shows that none of the tested hard cutoffs improves over using the render condition throughout inference, suggesting that simply disabling memory guidance at later denoising stages is suboptimal. The 100\% cutoff setting also remains slightly below full render conditioning, suggesting that introducing hard condition dropout during training provides no additional benefit in this setting. Timestep-aware gating further improves performance by retaining memory guidance throughout denoising while adaptively modulating its strength across timesteps, providing more flexible control.

\subsection{Fixed vs. Online 3D Memory}
\label{sec:appendix-fixed-online-memory}

To examine the practical difference between fixed and evolving memory configurations, we compare the camera-only baseline with fixed and online 3D memory under both short- and long-horizon evaluation. The fixed-memory variant constructs its 3D memory only from the initial observation and keeps it unchanged throughout the rollout. During training, its rendered condition is likewise always obtained from the initial frame. In contrast, the online-memory variant is trained with 1--4 randomly sampled source observations for memory reprojection and progressively incorporates newly generated observations during inference. Except for the first segment, where only the initial observation is available, we use four historical source observations for memory reprojection during rollout.

\begin{table}[h]
    \centering
    \caption{Comparison of fixed and online 3D memory. We report generation quality on both the first 81-frame segment and the full eight-segment (641-frame) rollout.}
    \label{tab:fixed-online-memory}
    \resizebox{\linewidth}{!}{
    \begin{tabular}{lcccccccccc}
        \toprule
        \multirow{2}{*}{Method}
        & \multicolumn{5}{c}{First Segment (81 frames)}
        & \multicolumn{5}{c}{8 Segments (641 frames)} \\
        \cmidrule(lr){2-6} \cmidrule(lr){7-11}
        & SSIM$\uparrow$ & PSNR$\uparrow$ & LPIPS$\downarrow$ & FID$\downarrow$ & $\mathrm{FID}_{\mathrm{cube}}\downarrow$
        & SSIM$\uparrow$ & PSNR$\uparrow$ & LPIPS$\downarrow$ & FID$\downarrow$ & $\mathrm{FID}_{\mathrm{cube}}\downarrow$ \\
        \midrule
        Wan2.1-Fun
        & 0.7181 & 20.88 & 0.1717 & 18.78 & 34.98
        & 0.5492 & 15.36 & 0.4220 & 24.82 & 45.21 \\
        + Fixed Memory
        & 0.7263 & 21.19 & 0.1691 & 18.87 & 35.51
        & 0.5280 & 14.76 & 0.4460 & 29.80 & 51.39 \\
        + Online Memory
        & \textbf{0.7357} & \textbf{21.41} & \textbf{0.1654} & \textbf{18.60} & \textbf{33.74}
        & \textbf{0.5806} & \textbf{15.95} & \textbf{0.4132} & 26.69 & \textbf{42.67} \\
        \bottomrule
    \end{tabular}}
\end{table}

As shown in Tab.~\ref{tab:fixed-online-memory}, fixed 3D memory provides modest improvements over the camera-only baseline in the first segment, indicating that reprojected geometric guidance is useful when the target views remain well covered by the initial observation. The online-memory variant performs slightly better at this stage. Since no memory update has yet occurred in the first segment, this difference is mainly associated with its multi-source training strategy, which exposes the model to more diverse reprojection conditions than the single-source training used by the fixed-memory variant.
 
The difference becomes more pronounced over extended rollout. After eight segments, the fixed-memory variant falls below the camera-only baseline across all reported metrics. As the camera explores regions beyond the initial observation, fixed-memory reprojections can provide less coverage, which may contribute to the observed degradation.

In contrast, the online-memory configuration continuously incorporates newly generated observations as exploration proceeds, allowing the memory to expand with the explored scene and provide guidance from more relevant historical views. This leads to substantially better long-horizon performance than the fixed-memory configuration and supports the use of an evolving 3D memory for extended panoramic wandering.

\subsection{PVMetric}
\label{sec:appendix-pvmetric}
To estimate depth for online 3D memory construction, we train PVMetric, a feed-forward panoramic video metric-depth estimator. It predicts depth at metric scale and is used during both training and inference of PVWander.

\subsubsection{Architecture Design}
\label{sec:pvmetric-architecture}

\begin{figure}[h]
  \centering
  \includegraphics[width=\linewidth]{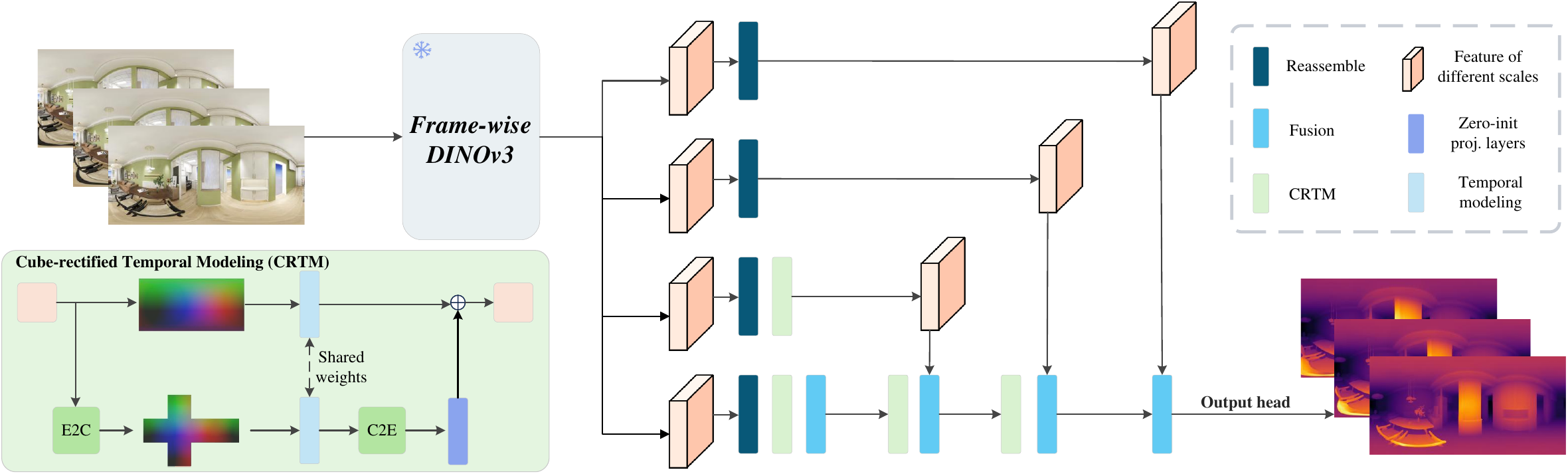}
  \caption{Architecture of PVMetric. The frame-wise DINOv3 backbone extracts multi-scale spatial features, while Cube-rectified Temporal Modeling (CRTM) modules perform temporal modeling at selected feature scales of the DPT decoder.}
  \label{fig:pvmetric-architecture}
\end{figure}

\paragraph{Overview.}
PVMetric extends the pretrained Depth Any Panoramas (DAP)~\citep{linDepthAnyPanoramas2025} image model into a feed-forward panoramic video metric-depth estimator. It retains the DINOv3 encoder~\citep{simeoniDINOv32025} and DPT multi-scale decoder of DAP, and introduces Cube-rectified Temporal Modeling (CRTM) modules at selected feature scales to capture cross-frame dependencies. This design builds on DAP's pretrained panoramic metric-depth representation while adapting its image-based decoder to video inputs. The overall architecture is summarized in Fig.~\ref{fig:pvmetric-architecture}.

\paragraph{Frame-wise Spatial Encoding.}
Given an ERP video $\mathbf{I}\in\mathbb{R}^{B\times T\times 3\times H\times W}$, we first fold the frame dimension into the batch dimension and process the resulting $(BT)\times 3\times H\times W$ images with a shared DINOv3 encoder. Following DAP, four intermediate encoder features are passed to the DPT decoder for
multi-scale spatial refinement. At this stage, each frame is encoded independently; cross-frame interactions are introduced by the temporal modules described below.

\paragraph{Multi-scale Temporal Modeling.}
Following Video Depth Anything~\citep{chenVideoDepthAnything2025}, we insert temporal layers at several lower-resolution positions of the DPT head to model cross-frame interactions while limiting the additional computational cost. Specifically, temporal modeling is introduced at four intermediate feature scales, whereas the higher-resolution features remain on the original frame-wise spatial pathway. For a feature at scale $s$, we gather the features at each spatial location across all frames into a temporal sequence:
\begin{equation}
  \mathbf{F}^{(s)}
  \in \mathbb{R}^{B\times C_s\times T\times H_s\times W_s}
  \xrightarrow{\operatorname{reshape}}
  \mathbf{F}_{\mathrm{seq}}^{(s)}
  \in \mathbb{R}^{(BH_sW_s)\times T\times C_s}.
  \label{eq:pvmetric-temporal-sequence}
\end{equation}
Self-attention is performed only along the $T$ dimension, whereas spatial interaction is handled by the DINOv3 encoder and DPT refinement blocks. Each temporal Transformer is incorporated as a residual update initialized to preserve the pretrained spatial pathway.

\paragraph{Cube-rectified Temporal Modeling.}
The latitude-dependent stretching of ERP causes smooth motion to manifest as increasingly nonlinear temporal dynamics toward the polar regions. We therefore incorporate Cube-rectified Temporal Modeling (CRTM) from PVDepth~\citep{songPVDepthPanoramicVideo} into the multi-scale temporal modules of PVMetric. At each temporal scale, CRTM processes the feature through parallel ERP and Cubemap branches:
\begin{equation}
\begin{aligned}
  \mathbf{Y}_{E}^{(s)}
    &= \mathcal{T}^{(s)}\!\left(\mathbf{F}^{(s)}\right), \\
  \mathbf{Y}_{C}^{(s)}
    &= \mathcal{P}_{C\rightarrow E}\!\left(
       \mathcal{T}^{(s)}\!\left(
       \mathcal{P}_{E\rightarrow C}\!\left(\mathbf{F}^{(s)}\right)
       \right)\right), \\
  \mathbf{F}_{\mathrm{out}}^{(s)}
    &= \mathbf{Y}_{E}^{(s)}
       + \mathcal{G}_{C}^{(s)}\!\left(\mathbf{Y}_{C}^{(s)}\right).
\end{aligned}
\label{eq:pvmetric-crtm}
\end{equation}
Here, $\mathcal{P}_{E\rightarrow C}$ and $\mathcal{P}_{C\rightarrow E}$ denote ERP-to-Cubemap and Cubemap-to-ERP projection, respectively. The six Cubemap faces are folded into the batch dimension and modeled independently along the temporal dimension. Within each feature scale, the ERP and Cubemap branches share the same temporal Transformer $\mathcal{T}^{(s)}$, while different scales use independent parameters. The fusion projection $\mathcal{G}_{C}^{(s)}$ is zero-initialized, allowing the Cubemap branch to be progressively learned as a residual correction to the ERP temporal features.

\paragraph{Metric Depth Prediction.}
The DPT decoder progressively fuses the temporally enhanced multi-scale features and predicts depth results for each input frame. During training, GT metric depth is clipped at $D_{\max}=100\,\mathrm{m}$ and divided by $D_{\max}$ to obtain values in $[0,1]$. To account for ERP's nonuniform spherical sampling, all loss terms are computed over valid pixels with cosine-latitude weighting. The training objective combines an absolute depth loss, a log-depth loss, and a temporal depth-difference loss~\citep{chenVideoDepthAnything2025}:
\begin{equation}
\mathcal{L} = \mathcal{L}_{\mathrm{L1}} + 0.2\mathcal{L}_{\log} + \mathcal{L}_{\mathrm{temp}}.
\label{eq:pvmetric-training-loss}
\end{equation}
Here, $\mathcal{L}_{\mathrm{L1}}$ and $\mathcal{L}_{\log}$ supervise the scaled metric depth in the linear and logarithmic domains, respectively, while $\mathcal{L}_{\mathrm{temp}}$ matches predicted and GT depth changes between adjacent frames. The temporal term is set to zero for single-frame samples. 
During inference, the prediction is multiplied by $D_{\max}$ to recover metric depth. For long-video inference, we follow the overlapping-window inference strategy of Video Depth Anything~\citep{chenVideoDepthAnything2025}, but omit the inter-window scale--shift alignment used for affine-invariant depth. The resulting metric depth sequence can be directly integrated into the online 3D memory for geometric reconstruction.

\subsubsection{Implementation Details}

\paragraph{Training Setup.}
We initialize PVMetric from the pretrained DAP model and freeze the DINOv3 encoder, while optimizing the DPT decoder and the four temporal modules. We use learning rates of $1\times10^{-5}$ and $5\times10^{-5}$ for the spatial and temporal layers, respectively. The model is optimized with AdamW using a weight decay of $10^{-2}$ and gradient clipping at 1.0 under BF16 mixed-precision training. 
We use effective batch sizes of 8 video clips and 160 images for the video and image modes, respectively, and train PVMetric for 150K optimization steps on four NVIDIA A100 GPUs, requiring approximately 636 GPU-hours. Both training and inference operate on ERP inputs resized to $1024\times512$.

\paragraph{Mixed Image--Video Training.}
Inspired by VDA~\citep{chenVideoDepthAnything2025}, we jointly train PVMetric with 32-frame video clips and single-frame images to learn temporal depth evolution while retaining the single-frame prior of DAP. At each training step, we select the video mode with probability 0.8 or the image mode with probability 0.2. The video mode uses PanoCARLA~\citep{songPVDepthPanoramicVideo}, PanoCity~\citep{guoPanoVGGTFeedForward3D2026}, and InteriorGS-OmniRoam~\citep{liuOmniRoamWorldWandering2026}, which together contain approximately 975K training frames. The image mode samples single frames from these video datasets and additionally includes Structured3D~\citep{ZhengStructured3D2020} and Omni360-Scene~\citep{geAirSim360PanoramicSimulation2025}; the latter two provide approximately 227K additional training images. Both modes share the same network and metric depth prediction target.

\subsubsection{Depth Estimation Evaluation}
\label{sec:pvmetric-evaluation}

\paragraph{PanoCARLA ablation and evaluation.}
To isolate the contribution of CRTM, we conduct a controlled ablation on the official PanoCARLA training and test splits. PVMetric variants with and without CRTM are trained only on PanoCARLA for 20K optimization steps with an effective batch size of 8. We additionally evaluate the selected architecture trained with the complete recipe alongside the frame-wise DAP baseline and the two controlled variants. We report absolute relative error (Abs Rel) and $\delta_1$, the fraction of predictions satisfying a depth ratio below $1.25$, under three video settings: 2 FPS / 50 frames, 10 FPS / 90 frames, and 20 FPS / 110 frames.

\begin{table}[h]
  \caption{Panoramic video metric-depth evaluation on PanoCARLA. The two 20K PVMetric variants are trained only on PanoCARLA under identical settings to isolate the effect of CRTM; PVMetric (full, 150K) denotes the final model trained with the complete recipe in Section~\ref{sec:appendix-pvmetric}.}
  \label{tab:pvmetric-panocarla}
  \centering
  \small
  \setlength{\tabcolsep}{3.2pt}
  \begin{tabular}{@{}lcccccc@{}}
    \toprule
    & \multicolumn{2}{c}{2 FPS / 50 frames}
    & \multicolumn{2}{c}{10 FPS / 90 frames}
    & \multicolumn{2}{c}{20 FPS / 110 frames} \\
    \cmidrule(lr){2-3}\cmidrule(lr){4-5}\cmidrule(l){6-7}
    Method
    & Abs Rel $\downarrow$ & $\delta_1\uparrow$
    & Abs Rel $\downarrow$ & $\delta_1\uparrow$
    & Abs Rel $\downarrow$ & $\delta_1\uparrow$ \\
    \midrule
    DAP
    & 0.314 & 0.619 & 0.297 & 0.660 & 0.294 & 0.664 \\
    PVMetric w/o CRTM (20K)
    & 0.199 & 0.702 & 0.180 & 0.749 & 0.178 & 0.752 \\
    PVMetric (20K)
    & 0.194 & 0.727 & \textbf{0.173} & 0.779
    & \textbf{0.171} & 0.785 \\
    \midrule
    PVMetric (full, 150K)
    & \textbf{0.190} & \textbf{0.794}
    & \textbf{0.173} & \textbf{0.832}
    & 0.172 & \textbf{0.836} \\
    \bottomrule
  \end{tabular}
\end{table}

Tab.~\ref{tab:pvmetric-panocarla} shows that all PVMetric variants outperform the frame-wise DAP baseline across the three video settings. Under the controlled 20K-step protocol, adding CRTM consistently improves both metrics, motivating its inclusion in the final architecture. The full 150K model further improves $\delta_1$ at all frame rates while maintaining comparable Abs Rel.

\paragraph{Zero-shot single-frame generalization.}
To examine whether the final PVMetric preserves the single-frame generalization ability inherited from DAP, we evaluate the final PVMetric model on Stanford2D3D~\citep{armeniJoint2D3DSemanticData2017} and Matterport3D~\citep{changMatterport3DLearningRGBD2017} without dataset-specific fine-tuning, following the evaluation protocol of DAP~\citep{linDepthAnyPanoramas2025}.

\begin{table}[h]
  \caption{Zero-shot single-frame metric-depth evaluation of DAP and PVMetric.}
  \label{tab:pvmetric-zeroshot}
  \centering
  \small
  \setlength{\tabcolsep}{4.2pt}
  \begin{tabular}{@{}lcccccc@{}}
    \toprule
    & \multicolumn{3}{c}{Stanford2D3D}
    & \multicolumn{3}{c}{Matterport3D} \\
    \cmidrule(lr){2-4}\cmidrule(l){5-7}
    Method
    & Abs Rel $\downarrow$ & RMSE $\downarrow$ & $\delta_1\uparrow$
    & Abs Rel $\downarrow$ & RMSE $\downarrow$ & $\delta_1\uparrow$ \\
    \midrule
    DAP
    & \textbf{0.1038} & \textbf{0.1760} & 0.9217
    & \textbf{0.1290} & \textbf{0.2589} & 0.8485 \\
    PVMetric
    & 0.1163 & 0.1892 & \textbf{0.9225}
    & 0.1304 & 0.2614 & \textbf{0.8490} \\
    \bottomrule
  \end{tabular}
\end{table}

Tab.~\ref{tab:pvmetric-zeroshot} shows that PVMetric achieves slightly higher $\delta_1$ on both datasets, whereas DAP retains lower Abs Rel and RMSE. 

\paragraph{InteriorGS-OmniRoam evaluation.}
Our main video generation experiments are conducted on InteriorGS-OmniRoam, where PVMetric provides the depth estimates used to construct the online 3D memory. We therefore evaluate PVMetric's metric-depth accuracy on the InteriorGS-OmniRoam test set.

\begin{table}[h]
  \caption{Metric-depth evaluation on the InteriorGS-OmniRoam test set.}
  \label{tab:pvmetric-interiorgs-omniroam}
  \centering
  \small
  \setlength{\tabcolsep}{8pt}
  \begin{tabular}{@{}lcc@{}}
    \toprule
    Method & Abs Rel $\downarrow$ & $\delta_1\uparrow$ \\
    \midrule
    DAP (zero-shot) & 0.1516 & 0.8462 \\
    PVMetric & \textbf{0.0623} & \textbf{0.9541} \\
    \bottomrule
  \end{tabular}
\end{table}

As shown in Tab.~\ref{tab:pvmetric-interiorgs-omniroam}, PVMetric achieves an Abs Rel of 0.0623 and a $\delta_1$ of 0.9541, supporting its use as the depth provider for online 3D memory in the downstream video generation experiments.

\subsection{Baseline Details}
\label{sec:appendix-baselines}
We provide implementation and evaluation details for the baselines in Tabs.~\ref{tab:main-comparison} and~\ref{tab:evoworld-comparison}.

\paragraph{EvoWorld.}
EvoWorld~\citep{wangEvoWorldEvolvingPanoramic2025} reprojects an evolving 3D point-cloud memory to guide panoramic video generation. We evaluate the released Unity\_Curve checkpoint at its native resolution of $1024\times576$ and resize its outputs to $1024\times512$ for comparison with SVD-based PVWander. In Tab.~\ref{tab:evoworld-comparison}, EvoWorld$^*$ denotes our reproduction. We compare this baseline with its PF+TAG variant under matched training and inference settings at $640\times320$. Appendix~\ref{sec:appendix-experimental-settings} gives the training schedule.

\paragraph{Wan2.1-Fun.}
We adapt \textit{Wan2.1-Fun-V1.1-1.3B-Control-Camera}~\citep{wanWanOpenAdvanced2025} to panoramic generation by replacing perspective rays with ERP spherical rays in its Pl\"ucker camera encoding, as detailed in Appendix~\ref{sec:appendix-erp-camera}. 
We fine-tune this camera-only baseline on InteriorGS-OmniRoam for a fair comparison.
At inference, it performs autoregressive rollout by using the last frame of each segment as the initial-frame condition for the next, together with the corresponding camera trajectory.

\paragraph{OmniRoam.}
OmniRoam~\citep{liuOmniRoamWorldWandering2026} adopts a two-stage preview-and-refinement pipeline. 
Given the full eight-segment trajectory, its preview model first generates sparse panoramic frames at $960\times480$ with a temporal stride of 8. 
These frames provide guidance across the full trajectory for subsequent segment-wise refinement at $1440\times720$. 
For comparison with other autoregressive methods, we also evaluate the released preview model without refinement, using each segment's last frame to initialize the next. 
We denote this variant as \textit{OmniRoam-P} in Tab.~\ref{tab:main-comparison}.

\paragraph{Matrix-3D.}
Matrix-3D~\citep{yangMatrix3DOmnidirectionalExplorable2025} estimates depth from the input panorama using MoGe~\citep{wangMoGeUnlockingAccurate2025} and constructs a scene mesh whose renders along the target trajectory provide geometric conditioning. 
For long-horizon rollout, we extend its single-segment generation autoregressively by using the last frame of each segment to initialize the next. At each segment, geometry is reconstructed from the current initial frame.

In addition to the released Matrix-3D model, we train two adapted variants on our dataset using the same ERP Wan2.1-Fun backbone as PVWander at $640\times320$, enabling a more controlled comparison. Matrix-3D$_{\mathrm{adapted}}^{\mathrm{MoGe}}$ follows the original depth-source protocol, using GT depth for geometric conditioning during training and MoGe-estimated depth during inference. The predicted depth of the initial frame is aligned to GT depth before projection. 
Matrix-3D$_{\mathrm{adapted}}^{\mathrm{PVMetric}}$ instead uses PVMetric predictions during both training and inference without GT alignment, matching the depth protocol used by PVWander.

As shown in Tab.~\ref{tab:main-comparison}, the PVMetric variant improves SSIM, PSNR, and LPIPS over the MoGe variant for both the first segment and the full rollout, supporting the benefit of maintaining a consistent depth source across training and inference. 
The MoGe variant achieves higher trajectory fidelity in the first segment, likely benefiting from the GT-aligned depth used for geometric projection. However, this advantage does not persist over the full rollout, where the PVMetric variant yields lower translation and rotation errors. This suggests that accurate depth alignment can benefit short-horizon geometric guidance, whereas maintaining a consistent depth-estimation pipeline becomes increasingly important under recursive long-horizon generation. Appendix~\ref{sec:appendix-depth-consistency} further analyzes the effect of training--inference depth consistency.

\paragraph{PanoWorld.}
PanoWorld~\citep{liPanoWorldRealworldPanoramic2026} incorporates panoramic camera conditioning and long-range memory for controllable panoramic generation. However, the released code and weights support only single-segment generation. We therefore extend it to long-horizon evaluation through autoregressive rollout, using the last frame of each segment as the initial-frame condition for the next.

\subsection{Camera Trajectory Evaluation}
\label{sec:appendix-camera-metrics}

\paragraph{Pose Estimation.}
Following the camera-controllability evaluation of WorldScore~\citep{duanWorldScoreUnifiedEvaluation2025}, we use ViPE~\citep{huangViPEVideoPose2025} to recover frame-wise camera poses from generated panoramic videos and compare them with the prescribed GT trajectories. The estimated and GT poses are converted to a common camera and world-coordinate convention.

\paragraph{Alignment.}
We left-multiply all ViPE rotations and positions by the GT first-frame rotation to express them in the GT world frame. Let $\hat{\mathbf{t}}_i$ denote the resulting estimated positions and $\mathbf{t}^{\mathrm{gt}}_i$ the GT positions for $N$ frames indexed by $i=0,\ldots,N-1$. We subtract each trajectory's first-frame position and estimate one global scale per sequence:
\begin{equation}
s^{*}=\arg\min_s\sum_{i=0}^{N-1}
\left\|\bigl(\mathbf{t}^{\mathrm{gt}}_i-\mathbf{t}^{\mathrm{gt}}_0\bigr)-s\bigl(\hat{\mathbf{t}}_i-\hat{\mathbf{t}}_0\bigr)\right\|_2.
\end{equation}

\paragraph{Metrics and Aggregation.}
TransErr measures the Euclidean distance between the scale-aligned camera positions, while RotErr measures the geodesic angular distance between camera rotations in degrees. TransErr uses the same unit as the GT camera positions. We first average each metric over all frames within a scene and then equally average the scene-level results over all scenes. Lower values indicate better trajectory fidelity.

\end{document}

%% file: math_commands.tex
\usepackage{amsmath,amsfonts,bm}

\def\eqref#1{equation~\ref{#1}}

\def\1{\bm{1}}

\DeclareMathAlphabet{\mathsfit}{\encodingdefault}{\sfdefault}{m}{sl}
\SetMathAlphabet{\mathsfit}{bold}{\encodingdefault}{\sfdefault}{bx}{n}

